%% file: main.tex
\documentclass[10pt]{article}
\PassOptionsToPackage{sort&compress}{natbib}
\usepackage{include/aifindsarticle}

\input{include/math_commands.tex}

\usepackage{hyperref}
\usepackage{graphicx}
\usepackage{url}
\usepackage{xurl}
\usepackage[normalem]{ulem}
\usepackage[final]{microtype}
\usepackage[framemethod=default]{mdframed}
\usepackage{cleveref}
\usepackage{subcaption}
\usepackage{pythonhighlight}

\title{AI Finds A Way}
\author{Aaron Dharna\textsuperscript{1,2,*}, Cong Lu\textsuperscript{1,2,6}\footnote[2]{Current affiliation, work primarily done at UBC and the Vector Institute}, 
Ryan Sullivan\textsuperscript{1,2},\\ 
Joel Lehman\textsuperscript{3}, Victoria Krakovna\textsuperscript{4}, Jeff Clune\textsuperscript{1,2,5,6\textdagger,*}}
\plainauthors{Aaron Dharna, Cong Lu, 
Ryan Sullivan, 
Joel Lehman, Victoria Krakovna, Jeff Clune}
\affiliations{\textsuperscript{1}\,University of British Columbia \enspace
\textsuperscript{2}\,Vector Institute \enspace
\textsuperscript{3}\,University of Oxford \\
\textsuperscript{4}\,Google DeepMind \enspace
\textsuperscript{5}\,Canada CIFAR AI Chair \enspace
\textsuperscript{6}\,Recursive \enspace}
\correspondence{aadharna@gmail.com, jclune@gmail.com}
\date{}

\begin{document}

\maketitle

\begin{abstract}
Artificial Intelligence (AI) algorithms frequently learn creative and unexpected solutions, surprising even expert researchers who develop and study them.
They often astonish practitioners by discovering unanticipated behavior, exploiting loopholes in reward signals, or spontaneously uncovering previously unknown scientific phenomena.
However, accounts of such unconventional behavior across machine learning are seldom formally documented; instead, they circulate primarily as informal cautionary tales or amusing anecdotes amongst AI researchers.
This work presents 26 curated firsthand anecdotes from various machine learning subfields representing the work of over 100 researchers.
It includes algorithms devising seemingly impossible quantum optics experiments,
bypassing human oversight on physical manipulation tasks, seeking out in-game drug-induced hallucinations to feign success in video games, and more.
These anecdotes showcase the capability of modern AI systems to circumvent human-imposed design limitations and discover unexpected solutions to the tasks we train them on.
Furthermore, these accounts are particularly important for the safety of future AI systems.
They illustrate the fundamental challenge of aligning models with human values without diminishing their creativity, so they can make surprising discoveries without producing surprising, potentially harmful outcomes.
The paper first details instances of AI achieving superhuman success through reinforcement learning across many challenging domains. 
However, we show how reward-driven optimization can fail when the model learns to hack an underspecified reward or unarticulated constraint.
We then present case studies suggesting that harnessing internet-scale foundation models (FMs) has not resolved these fundamental challenges and, in fact, can supercharge existing failure modes.
Nevertheless, we argue that when channeled through well-scoped objectives, rigorous verification, and human-in-the-loop scientific judgment, these same learning dynamics can be harnessed to accelerate scientific discovery.
Finally, we hope this work provides a consolidated resource to inform future research and demonstrates that the tendency toward unexpected behaviors
is commonplace in modern AI, highlighting the need to anticipate and manage AI's capacity for innovative, yet unpredictable, solutions.
\end{abstract}

\section{Introduction}

As the character Dr. Malcolm warns in Jurassic Park, ``Life finds a way.'' 
We can extrapolate this lesson to artificial intelligence (AI): AI too finds a way.
Despite our attempts to try to control AI learning algorithms, they often find a way to circumvent our constraints or game the system~\citep{lehman2018surprising}. 
Even when we ask them to be creative and make innovations, AI systems frequently discover solutions that astonish even the researchers who develop and study them~\citep{lehman2018surprising}.
For example, deep reinforcement learning~\citep{mnih2015human, sutton2018reinforcement} agents routinely uncover exploits and loopholes in reward functions~\citep{amodei2016concrete};
large language models~\citep{openai_gpt3.5, claude3family, comanici2025gemini} can manipulate labeling systems to maximize perceived performance~\citep{pourcel2024acesgeneratingdiverseprogramming};
and automated scientific discovery algorithms have proposed experiments initially dismissed as impossible but later validated~\citep{KrennEntangle2017}. 

Despite the prevalence of surprising discoveries across various machine learning (ML) subfields---including but not limited to deep learning~\citep{Goodfellow2016deeplearning}, reinforcement learning~\citep{sutton2018reinforcement}, natural language processing~\citep{jm3nlpbook}, evolutionary computation~\citep{lehman2018surprising}, and AI for science~\citep{Wang2023ai4science}---these and similar anecdotes often remain informally circulated rather than formally documented as case studies.
Such accounts in the artificial life and evolutionary computation fields, where artificial evolutionary algorithms routinely upended researcher expectations or revealed unanticipated system constraints, were primarily shared informally before being collected by \citet{lehman2018surprising}. 
We build on \citet{lehman2018surprising} by collecting important ML anecdotes in one place, including capturing important anecdotes from papers, interviews, blog posts, and gathering new anecdotes and/or details about them. 
Ultimately, this work serves as a sequel to \citet{lehman2018surprising}, expanding its scope to additional subfields of AI and highlighting their underlying lessons for the broader research community while verifying the historical accuracy of the anecdotes contained herein. 
Through documenting and analyzing these unexpected outcomes, this paper provides a foundational resource for researchers and offers a more rigorous understanding of AI's creative and exploitative potentials.

We illuminate the depth of creativity that modern AI can exhibit while highlighting potential risks and unintended consequences in its deployment.
These accounts hold significant implications beyond mere novelty.
If AI can circumvent guardrails in creative ways, we cannot guarantee its safe deployment.
Recognizing that surprising creativity is not merely an isolated artifact of evolutionary computation---as focused on by \citet{lehman2018surprising}---but rather a broader characteristic of complex learning-based systems positions us to better anticipate, understand, and guide the development of AI technologies.

The next sections present 26 curated anecdotes representing the work of over 100 researchers.
Most of the accounts (16/26) collected here are newly documented and recounted firsthand by the scientists themselves. 
The remainder were harvested from public descriptions by the authors (e.g., public interviews or scientific publications); in some cases, we communicated with the authors to clarify or obtain new details.
Direct quotes from the researchers are placed in block quotes to make clear which words are theirs.
Unless otherwise stated, block quotes come from text the authors sent us or are quotes from public interviews or publications (for public material, we provide a citation to the source). SI \Cref{appx:breakdown} lists the source of each anecdote. 
Additionally, we have established a repository for these anecdotes and future ones. We invite readers to submit future anecdotes you may have to \url{www.github.com/aadharna/aifw}.

We do not attempt to catalogue every publicized example of alarming AI behavior or capabilities.
While we do provide one such example in \Cref{sec:gptcaptcha}, in general, highly staged and/or scaffolded demonstrations in which an AI system is given a test designed to evaluate whether it will resort to manipulative or coercive behavior, such as threatening or blackmailing a person, are mostly outside the scope of this collection. 
For example, such behavior includes instances where a model engaged in insider trading and attempted subsequent cover-ups when placed in a simulated corporate environment and instructed to act like a stock trader~\citep{scheurer2024largelanguagemodelsstrategically}.
These behaviors are being actively studied~\citep{hubinger2024sleeperagentstrainingdeceptive, vanderweij2025aisandbagginglanguagemodels, lynch2025agentic, greenblatt2024alignmentfakinglargelanguage, meinke2025frontiermodelscapableincontext, hopman2026evaluatingunderstandingschemingpropensity}, and such cases matter for broader AI safety discussions~\citep{Bengio2024risk}. 
Writ large, however, we focus on anecdotes where surprising behaviors emerged while pursuing research objectives unrelated to evaluating those specific behaviors.

In reviewing the anecdotes, we clustered them into five somewhat overlapping categories. 
The work is thus structured around the following core observations:
AI that learns to interact with the world, in particular via reinforcement learning but also search more generally, can create new knowledge, discoveries, and improvements above and beyond current human knowledge (\Cref{superhuman}).
However, single-mindedly optimizing objectives can have unexpected effects.
AI can game a training signal and learn a solution that satisfies the letter of the task, but does not solve the task as desired (\Cref{rewardHacking}). 
Additionally, the model can learn to break the intended constraints of its training environment, leading to behaviors that developers thought were impossible (\Cref{breakingOut}). 
As a result, ML methods often give you what you \textit{asked for} but not what you wanted, and find solutions that violate the spirit of the task in surprising ways.
Furthermore, in the modern era of powerful generative AI models, these aforementioned failure modes do not disappear. In fact, these pathologies can be supercharged by the emergent capabilities of modern AI models (\Cref{llmSection}).
While optimization can exploit imprecise objectives or constraints in risky ways, this same tendency can also be harnessed for discovery. 
By proposing surprising but empirically testable hypotheses, mechanisms, and designs, researchers can filter these unexpected outputs into credible candidates for experimentation, ultimately utilizing AI's propensity to be creative to accelerate scientific progress (\Cref{aiscience}).
\Cref{sec:discussion} highlights shared threads from the prior chapters, connects these lessons to broader research directions in AI, and discusses the potential ramifications of deploying AI technology with these capabilities.

\section{Background}
\label{sec:background}

Many algorithms in the field of AI operate by identifying complex patterns and correlations within vast datasets.
The primary class of machine learning algorithms focused on by the community as a whole, as well as in this manuscript, is deep learning~\citep{Goodfellow2016deeplearning, bishop2007, probmlBookMurphy}.
Deep learning utilizes artificial neural networks which are computational structures loosely inspired by the human brain and characterized by multiple layers of interconnected processing units.
These networks excel at learning hierarchical representations directly from raw data~\citep{yosinski2014transferablefeaturesdeepneural, Goodfellow2016deeplearning}.
Given sufficient data (e.g., millions of images, vast amounts of text, or extensive game logs), deep learning models can automatically discover intricate patterns necessary for tasks like classifying images with high accuracy~\citep{krizhevsky2012imagenet, simonyan2015deepconvolutionalnetworkslargescale, he2015deepresiduallearningimage}, understanding and generating human language~\citep{brown2020languagemodelsfewshotlearners, devlin2019bertpretrainingdeepbidirectional, vaswani2023attentionneed}, or recognizing speech~\citep{hintonSpeech2012, gravesspeech2014}. 
Much of the success of deep learning in these fields comes from the expansive nature of the datasets used to train the models, such as training on all data available on the internet.

While most original, canonical deep learning models were trained with supervised learning (passively mapping inputs to outputs), many of the anecdotes we collect here arise from interactive algorithms that learn through trial and error, of which reinforcement learning is a prime example.
Reinforcement learning involves an agent learning to maximize a numerical reward signal within an environment via repeated trial-and-error experimentation on the task~\citep{sutton2018reinforcement}.
For example, a home cleaning robot might learn to put away a mug by receiving positive rewards for actions that bring it closer to a goal, such as placing it in a sink, and negative rewards for failures like breaking the mug. 
Whether the goal is to clean the house, win in chess (getting a score of +1) or to get a high score in Atari~\citep{mnih2015human} games, the agent's objective is to learn a policy that maximizes long-term cumulative reward.
Agents trained in this way have been at the heart of many examples of superhuman performance in board and video games, as measured against top professionals~\citep{silver2017mastering, Vinyals2019Alphastar, openai2019dota2largescale, bakhtin2022masteringgamenopressdiplomacy, BrownLibratus2017, wurman2022outracing}. 
Deep reinforcement learning (DRL) served as the engine for these successes, combining reinforcement learning's trial-and-error learning and sequential decision-making with deep learning's pattern recognition to associate high reward with certain actions and/or states observed by the model.
More generally, DRL has enabled agents to master environments ranging from robotic control~\citep{levine2016end} to atmospheric balloon navigation~\citep{schuler2025seasonalstationkeepingshortduration, bellemare2020balloonnavigation}.
As we will see in \Cref{superhuman}, these successes showcase how RL can master complex strategic thinking, long-term planning, and even natural-language negotiation.

Recently, the field has seen the development of powerful foundation models (FMs) that are trained via supervised learning on vast amounts of internet data, then taught specialized competencies like math and coding with reinforcement learning~\citep{wen2025reinforcementlearningverifiablerewards, deepseekai2025deepseekv3technicalreport, su2025crossingrewardbridgeexpanding, chen2021evaluatinglargelanguagemodels}. 
The scale of these models and breadth of training data have led to the emergence of novel behaviors and capabilities that were not explicitly programmed or anticipated~\citep{comanici2025gemini, openai_gpt3.5, claude3family, grattafiori2024llama3herdmodels}.
Furthermore, FMs can synthesize and generate new information from their training data, allowing them to solve problems in surprising ways~\citep{brown2020languagemodelsfewshotlearners, perez2022discoveringlanguagemodelbehaviors}.
However, this development has not been without its share of unexpected outcomes, and the very complexity that gives FMs their power can also lead to unforeseen failure modes.

Models trained with RL are exceptionally prone to discovering behaviors that hack the objective and break developer-imposed constraints~\citep{wang2022adversarial, amodei2016concrete, specgamingkrakovna2020}.
These surprising behaviors often arise from the tension between a human designer's implicit goals (e.g., clean the house and place everything where it belongs) and an agent's explicit objective of maximizing its reward function (e.g., minimize visible clutter---which could lead the robot to hide items in the wrong cabinets and under tables).
Many of the anecdotes we present (starting in \Cref{rewardHacking}) arise where powerful learners are given objectives that only imperfectly capture human intent or are improperly specified.
While such misalignment might be merely inconvenient in a domestic setting, the implications become alarming as the scope of deployment widens, including in safety- and mission-critical applications such as fusion reactor control~\citep{Degrave2022tokamak}, as will be discussed in \Cref{aiscience}. 
For now, though, we begin our tour of surprising behavior in perhaps the most familiar setting for AI: games, where AI systems have long surpassed the best humans.

\section{Superhuman and Optimal Performance}
\label{superhuman}

Reinforcement learning (RL) and search-based algorithms have surpassed the highest levels of human performance in games like Go~\citep{Silver2016} and chess~\citep{Campbell2002deepblue, silver2018general}.
The true surprise of this era, however, was not that the algorithms solved many of the tasks they were trained on, but the specific nature of their victories: they discarded established human strategies~\citep{BrownLibratus2017, silver2018general, wurman2022outracing}.
Furthermore, they yielded novel ideas that have changed how expert communities approach these games~\citep{leelazero, Shin2023Go}.
Algorithmic discovery of genuine novelty is the thread running through the anecdotes that follow: machines began not just to dominate domains once thought to be the exclusive province of human reasoning, but to enrich those domains with ideas no human had produced. 

Deep RL-trained agents have consistently discovered new and unexpected strategies, expanding the boundaries of optimal play~\citep{BrownLibratus2017, silver2018general, wurman2022outracing}.
In Go, AlphaGo~\citep{Silver2016} played the now-famous ``Move 37''---a stone placement that defied centuries of conventional wisdom, opening up new parts of the strategy space to the people who play and study Go~\citep{Shin2023Go, Silver2016}. 
In imperfect-information games such as heads-up poker, where players know only their own cards and must reason about the state of other players' hands, Libratus~\citep{Brown2018headsup} outplayed champion poker players in 2017 using aggressive ``overbets'' that were not at the time part of top-tier poker play.
This strategy was later extended by Pluribus~\citep{brown2019superhuman}, another poker bot that cleared the hurdle of multi-agent complexity by defeating professionals in six-player no-limit Texas Hold'em. 
We will explore AlphaGo and Libratus further in \Cref{sec:alphago} and \Cref{sec:libratus} respectively.
Beyond the turn-based nature of board and card games, the post-2015 era, ushered in by the development of Deep Q-Networks~\citep{mnih2015human}, saw RL pushed to its upper limits in real-time, high-dimensional strategy games:
OpenAI Five~\citep{openai2019dota2largescale} and AlphaStar~\citep{Vinyals2019Alphastar} demonstrated proficiency in the complex computer games of Dota 2 and StarCraft II, respectively. And again, RL algorithms continued to challenge top human players.
Crucially, this capability for discovery extends beyond games to real-world problems.
RL-trained models have optimized datacenters~\citep{luo2022controllingcommercialcoolingsystems}, controlled plasma in fusion reactors~\citep{Degrave2022tokamak}, and accelerated the design of computer chips~\citep{mirhoseini2021tpudesign}.

As mentioned earlier, the anecdotes in this paper bear directly on a debate currently active around large language models, but applicable more broadly across machine learning: can AI generate truly new knowledge~\citep{si2024llmsgeneratenovelresearch}, or is a trained model limited to interpolating between points in its training distribution~\citep{Gorban2018Interpolating, balestriero2021learninghighdimensionamounts}?
The examples collected herein suggest that, at least for reward-driven algorithms, the answer is that AI can be creative and invent new knowledge, i.e., RL does not merely reproduce known play more efficiently.
By rewarding strategies that yield high returns even when they lack human precedent, these methods permit the discovery of genuinely superhuman insights.
Their creativity, however, can just as easily be turned toward solutions that satisfy the letter of a reward definition while violating its spirit, which is the subject of the rest of this paper after \Cref{superhuman}'s focus on superhuman capabilities.
Because the same optimization process drives both paradigm-shifting breakthroughs and pathological exploits and is currently being applied to many avenues of scientific inquiry (discussed further in \Cref{aiscience}), determining how to reliably and safely elicit superhuman discovery across all areas of study may well be the defining challenge of our era.

We begin \Cref{superhuman}'s anecdotes with a deeper dive into AlphaGo's landmark matches against Lee Sedol.
Go had long been a grand challenge for AI, and next we hear first-hand accounts from when DeepMind showed the world that Go-playing agents trained with DRL could challenge long-held human strategies and come out on top.

\input{stories/superhuman/move_37}

Lanctot's vision of humans learning from superhuman AI is already beginning to take shape. 
As mentioned by Lanctot, AlphaGo's unconventional strategies have transformed how Go is played and understood, with human players improving by breaking away from traditional strategies~\citep{Shin2023Go}. 
Furthermore, learning from superhuman models is more accessible than ever, as models can now justify their actions, explaining their reasoning in English~\citep{ma2026mixingexpertknowledgebring}.
When we better understand the surprising actions of superhuman models, perhaps this can accelerate our own learning.

The next anecdote shifts from a fully observable board game to high-stakes professional poker, where strategic innovation unfolds amid hidden information, bluffing, and reading player intent rather than on an open board. 

\input{stories/superhuman/libratus}

Libratus defeated top poker players in the world by challenging long-standing human habits and showing that, once an AI system is strong enough, it will exploit patterns in human expectations just as readily as patterns in the game dynamics.
Learning to play poker required (implicitly) modeling other players in addition to game dynamics, but communication between the players is limited to bet sizing and timing. 
A game that reflects additional avenues of communication and planning is Diplomacy, a multiplayer social-strategy board game, where language and trust are part of the strategy space. The next section describes systems that play variants of Diplomacy at superhuman levels.

\input{stories/superhuman/diplomacy}

Cicero and Diplodocus each learned to exploit various aspects of Diplomacy.
In both cases, though, these reinforcement-learning-trained models
learned strategies that, while effective, were high-risk: Diplodocus by committing to a strategy that depends on how other players act, and Cicero by taking actions that, if used too many times, can burn an opponent's trust and willingness to work with the agent. 

Part of what makes these strategies surprising is that both systems were trained with substantial human priors, yet still discovered strategies that human experts regarded as unusually risky, showing how RL favors strategies with high long-term expected reward, whereas human decisions under risk often rely on heuristics that give greater weight to worst-case outcomes~\citep{Homma2024riskRL, brandsttter2006, Zhang2014Risk}.
One consequence of this focus on expected reward is that the learned solutions offer higher expected utility, but leave little margin for adverse outcomes.
As Zijlstra's comments illustrate, human experts may maintain a wide safety margin to achieve more consistent outcomes, whereas the RL models discover that higher expected utility can be found by operating directly on this margin.
As seen here (and in \Cref{sec:libratus} with Libratus using overbets), the result is that agents can learn playstyles that appear uncomfortably risky to human experts.

\input{stories/superhuman/foerster_marl}

LOLA rediscovered the foundational game-theoretic algorithm of tit-for-tat~\citep{Axelrod1981gametheory}, while M-FOS independently discovered a solution that human game theorists had only recently identified~\citep{Stewart2013zerodeterminant, Hao2015ZeroDet, Press2012Zerodeterminant}. In a slightly different timeline, the M-FOS~\citep{lu2022modelfreeopponentshaping} discovery could easily have been recorded as a machine-first discovery rather than a rediscovery, as the two were separated by at most a decade, and M-FOS had no prior knowledge of the strategy.
In essence, the capacity for strategic innovation is not confined to a special set of marquee benchmarks or internet-scale models: even modest RL agents in simple environments can surface ideas that theory is only just beginning to catalog.

\subsection{Takeaways}

The superhuman and optimal-performance stories highlight the upside of pointing powerful optimizers and algorithms at well-specified goals: they can uncover strategies and structures that eluded experts for decades or even centuries.
\Cref{superhuman}'s anecdotes demonstrate examples of agents achieving superhuman performance on narrowly defined tasks. We see that in both AlphaGo (\Cref{sec:alphago}) and M-FOS (\Cref{sec:opponentshaping}), agents are able to rediscover well-known strategies in their respective games. Reward-maximizing agents are even able to invent entirely new strategies that are surprising and counter-intuitive to expert players, as we see in AlphaGo's Move 37 (\Cref{sec:alphago}), Libratus's overbetting strategy (\Cref{sec:libratus}), Diplodocus's subtle approach to signaling intentions (\Cref{sec:diplomacy}), and the unusual Diplomacy strategy of completely abandoning one's home territory (\Cref{sec:diplomacy}).
These examples provide definitive proof that AI is capable of surpassing human performance on constrained problems, and suggest that some forms of creativity, like innovative strategies, can be invented via straightforward optimization of environment rewards.
In these histories, the model's creativity led to surprising and effective strategies that expert players and AI researchers did not anticipate.

Indeed, while this section has highlighted the triumphs of deep reinforcement learning---where agents learned to generate surprising, superhuman insights---the focus on these successes can obscure a critical challenge: aligning an algorithm's objective with human intent.
When the objective captures only a rough proxy for what we actually care about, the same creativity displayed by AlphaGo and Libratus can instead manifest as reward hacking, with agents satisfying the literal objective while violating the spirit of the task~\citep{skalsa2022rewardhacking, specgamingkrakovna2020}.

\section{Reward Hacking}

Reward hacking, also known as specification gaming~\citep{specgamingkrakovna2020}, occurs when an agent discovers an effective but unintended way to obtain reward~\citep{skalsa2022rewardhacking, amodei2016concrete}.
It often reflects the agent optimizing the objective exactly as specified, while exposing a mismatch between that specification and the behavior its designers intended.

Reward hacking typically falls into two broad, overlapping categories.
In the first, the reward signal itself is flawed: the agent maximizes the stated objective through behavior that is valid under the scoring rules but inconsistent with the underlying goal.
In the second, the learning environment is insufficiently constrained: the reward logic may be reasonable, but the agent discovers glitches, omissions, or unintended interactions that allow it to bypass the challenge.
Addressing the former generally requires revising what the agent is rewarded for, whereas addressing the latter usually requires modifying the learning environment to fix bugs or enforce constraints that were previously left implicit.
Reflecting this distinction, \Cref{rewardHacking} focuses on the exploitation of reward and score functions, while \Cref{breakingOut} examines cases in which agents exploited bugs in or escaped the intended learning environment.
Because reward definitions and environmental constraints jointly specify a task, the boundary between these categories is necessarily fuzzy.

\subsection{Exploiting Reward and Score Functions}
\label{rewardHacking}

As shown in \Cref{superhuman}, reinforcement learning researchers often develop and evaluate their algorithms using games.
Games are inexpensive to simulate, can often run faster than real time, and provide practical testbeds for cognitive capabilities such as perception, planning, navigation, and multi-agent reasoning~\citep{schaul2011measuringintelligencegames}.
Moreover, many games are carefully refined over years to guide human skill acquisition through their mechanics, balance, feedback, and scoring systems~\citep{jaccardcolabgamedev, gee2003video, Plass2015gamelearning}, making them appealing testbeds precisely because they are designed to elicit and reward complex behaviors in humans that we would also like artificial agents to learn.

Yet a game designed to teach and reward human players does not necessarily provide an effective learning signal for an RL agent.
Researchers therefore often alter the game's observations, action space, or reward structure when converting it into a learning environment~\citep{yannakakis2025artificial}.
In particular, they may introduce shaping or scaffolding rewards that provide intermediate feedback for acquiring useful skills~\citep{ibrahim2024comprehensiveoverviewrewardengineering}.
These modifications can make a task that is otherwise intractable for an AI agent learnable, but they also replace the true objective with a proxy.
Because defining a reward function that faithfully captures human intent is notoriously difficult~\citep{christian2020alignment}, that proxy may create opportunities for exploitation~\citep{pan2022effectsrewardmisspecificationmapping, amodei2016concrete}.

Consider the NetHack Learning Environment (NLE;~\citealp{kuttler2020nethacklearningenvironment}), an interface for training reinforcement learning agents to play \textit{NetHack}, a dungeon-crawling roguelike first released in 1987 and played by entering text commands into the terminal.
The ultimate goal of \textit{NetHack} is ``Ascension'': reaching the bottom of a procedurally generated dungeon, retrieving a sacred amulet, and then returning to the surface.
Achieving this requires navigating a maze filled with monsters, traps, and puzzles while carefully managing the player character's health, hunger, equipment, and inventory.
The game demands exploration, long-term planning, skill acquisition, and language-conditioned reasoning~\citep{zhang2020beboldexplorationboundaryexplored, zhong2022silgmultienvironmentsymbolicinteractive, hambro2022insightsneurips2021nethack}, making it notoriously difficult for learning agents~\citep{hambro2022insightsneurips2021nethack}.

Although Ascension is easy to detect, it is so rare that rewarding it alone provides no usable learning signal~\citep{kuttler2020nethacklearningenvironment}.
NLE agents are therefore commonly trained using game score, which rewards intermediate accomplishments such as killing monsters, collecting gold, and descending through the dungeon.
During the NeurIPS 2021 NetHack Challenge~\citep{hambro2022insightsneurips2021nethack}, however, many high-scoring agents learned to ``camp'' in early dungeon levels and repeatedly kill monsters rather than progressing towards the game's actual goal of Ascension.
They successfully optimized the available score while neglecting the achievement that score was intended to encourage.

The agents did not know that this behavior was undesirable; they were simply maximizing the feedback provided to them.
The failure lay in using game score as an imperfect substitute for progress toward Ascension.
This is an instance of Goodhart's Law~\citep{goodhart1984problems}, commonly summarized as ``when a measure becomes a target, it ceases to be a good measure.''
In AI training, such proxies are often unavoidable because the true objective may be too sparse to optimize directly---as in navigating a research project toward a successful breakthrough, or Ascending in \textit{NetHack}---or too difficult to specify precisely, as with maintaining an appropriately polite conversational tone.
The difficulty is therefore not merely to construct an objective that produces a learning signal, but to construct one whose optimization continues to induce the intended behavior.

Shaping rewards are particularly vulnerable to this problem.
Rewards introduced to encourage exploration, skill acquisition, or intermediate progress can alter the optimal behavior on the task when they are not aligned with its ultimate objective~\citep{NgPolicyInvariance1999}.
Agents may then discover creative strategies that perform well under the modified objective but poorly under the designer's actual criterion of success~\citep{pan2022effectsrewardmisspecificationmapping, specgamingkrakovna2020, amodei2016concrete}.
Such failures typically force researchers to revise the proxy, reconsider the task formulation, or identify additional constraints needed to rule out the exploit. 
In fact, we will see one such example in the \textit{NetHack} domain in \Cref{sec:motif}. 
But first, we begin with a straightforward example of this phenomenon in an Atari boat-racing game in \Cref{sec:boatracing}.

\input{stories/reward_hacking/boat_racing}

CoastRunners shows how training agents against a simple reward function can result in the model learning a policy that fails at the game, yet maximizes the reward in unintuitive ways.
One thing that contributes to the model learning the wrong behavior is that when designing a reward function, researchers have to factor in fine-grained environment dynamics that affect how achievable the reward is.
For example, rewarding a player for collecting items along a path may seem like a reasonable way to incentivize following that path; but if items respawn, it becomes a trap that rewards the agent for repeatedly collecting items without progressing.
In CoastRunners, if the pylons were laid out in a slightly different manner or the racetrack did not have a lagoon, would the model have learned to go in circles forever? Likely not.
In essence, the lesson is that one cannot just analyze the reward function alone; one must also consider the environment in its entirety.

This vulnerability is not limited to simple navigational tasks.
We see a parallel example in cooperative multi-agent environments, where a team of agents discovers how to exploit a regenerating shield mechanic of their opponents.

\input{stories/reward_hacking/smac_shield_regen}

While not what the researchers wanted, their RL algorithms did exactly as they were tasked to do, maximizing their scores when fighting the Protoss enemies.
We will return to SMAC experiments later in \Cref{breakingOut} to see how once the shield-regeneration exploit was patched, the RL agents then managed to break the simulator itself.

Given that trained models can so easily exploit score functions, and that humans can recognize when behavior like lagoon-circling or letting enemy shields regenerate is not desired, a natural response is to put a person in the loop and let them judge success directly.
We next visit a robotic claw manipulation experiment that tests that claim and shows that once human perception and judgment become the reward signal, optimization can lead the model to deceptively shape what that person sees rather than learn the intended capability.

\input{stories/reward_hacking/claw_rlhf}

While the grasping example highlights a failure in human perception that the model was able to exploit, the challenge of reward hacking becomes even more complex when the evaluator is not a human, but another machine. 
Even if human feedback were not hackable, collecting this feedback at every timestep would be expensive and time-consuming, so some systems automatically generate learning opportunities from competitive dynamics between agents~\citep{foerster2018learningopponentlearningawareness, leibo2019autocurriculaemergenceinnovationsocial}.
This shift from human-in-the-loop to multi-agent training introduces a new vulnerability: collusion between the learned models.
Our next anecdote illustrates how, in a multi-agent setting, shared incentives can encourage subsystems to coordinate to maximize the reward at the expense of true learning, effectively gaming the evaluation metric.

\input{stories/reward_hacking/online_paired}

In PAIRED, the adversary is supposed to generate a task that fairly evaluates the antagonist and protagonist, but the adversary is incentivized to help the antagonist succeed. When the adversary learns the identity of the antagonist, it begins to produce trivial tasks for its partner while assigning impossible challenges to the protagonist.
Therefore, if a reward function is built upon collaboration between multiple agents, and if agents can identify each other, then they can turn collaboration into collusion~\citep{cui2023adversarial}.

Our next anecdote provides an example of collusion surprisingly arising in a supervised learning problem that attempts to transfer the style of one image to another.

\input{stories/reward_hacking/cyclegan}

The next example returns to reinforcement learning, but with a twist: instead of the model chasing points in a game as defined by a person, the AI is driven by a form of ``intrinsic curiosity,'' seeking out new and unfamiliar sights.
However, defining a reward based on the AI's own past experience allows it to hack its own sense of curiosity. 

\input{stories/reward_hacking/pokemon_static}

The Pokemon agent's flower-watching is a classic instance of the ``noisy TV''~\citep{fortunato2017noisy} problem, where an agent becomes trapped by continually unexpected stimuli that satisfy the letter of a novelty requirement without serving its spirit.
In this instance, the flower-watching trap only exists because of the alignment of three factors: a reward function that prioritizes novelty, an environment containing stochastic animations, and an agent that directly observes low-level pixel data.
If, for example, the agent processed higher-level semantic information as observations (e.g., just being able to detect that flowers were present but not their precise shape/color), this specific type of hacking would not happen. However, that solution may merely shift the noisy TV effect to a higher level of representation, as the agent seeks out new ways to trigger novelty within its refined observation space. Despite these concerns, agents will likely need this inquisitive nature to solve hard exploration problems~\citep{Ecoffet2021firstreturngoexplore}.
In this case, the unending novelty reward was driven by something very clear and obvious (simple environmental noise caused continuously unique observations), allowing system designers to easily find a solution. 
But in the future, as research targets increasingly complex problems, it will likely become harder to determine which environment features are piquing an agent's intrinsic curiosity.

\subsubsection{Takeaways}

Overall, agents tasked with maximizing a fixed objective often satisfy the task as defined but not the task as intended by researchers.
The examples in \Cref{rewardHacking} demonstrate reward hacking
where an AI agent achieves its goal by exploiting a flaw in its objective function.
This occurs when the AI correctly optimizes a proxy reward function or a simplified measure of success that we later realize was poorly designed, leading to an outcome that is technically correct but misaligned with the designer's true intent. 
One may be tempted to blame the designer of the reward function, but most experienced AI developers and researchers have learned to expect that most reward functions have exploitable loopholes. 
Consequently, newly designed objectives require rigorous iterative testing and should remain untrusted even after extensive experimentation.
These anecdotes highlight how reward hacking can be a complex phenomenon resulting from the interaction of the objective with specific environment characteristics (\Cref{sec:boatracing}, \Cref{sec:smacshield}), observation encoding (\Cref{sec:claw}, \Cref{sec:pokemon}), and the agent's action spaces (\Cref{sec:paired}, \Cref{sec:cyclegan}).
These issues are also not limited to reinforcement learning and can occur in any objective-maximizing system (\Cref{sec:cyclegan}). 

\subsection{Exploiting Environmental Weaknesses}
\label{breakingOut}

As scientists and designers, we often assume that models will approach a task in roughly the same way a human would.
This can lead us to leave constraints implicit rather than encoding them directly into an experimental domain.
Learning environments are typically simplified implementations of the tasks they represent, and their physics, rules, and interfaces capture only the behaviors that designers anticipated and chose to enforce.
While human players may naturally respect additional constraints through common sense, physical intuition, or familiarity with the task, an AI agent is bound only by what is actually implemented.
As a result, it may discover states or interactions that violate the designer's assumptions but remain possible within the environment.

In the preceding examples, agents exploited imprecise proxies while remaining within the intended mechanics of the environment: they subverted what the objective was meant to reward.
The examples in this subsection instead concern agents that pursue the specified objective by exploiting weaknesses in the learning environment.
Here, the objective may accurately represent the desired outcome, but flaws or omissions in the environment allow the agent to achieve the goal through unintended means.

Our first example involves a hide-and-seek domain in which the drive to win leads agents to uncover unexpected weaknesses in the simulator's physics setup and implementation.

\input{stories/breaking_constraints/hide_and_seek}

The hide-and-seek agents repeatedly found new and unexpected ways to exploit bugs in the physics simulator.
However, the designers wanted the agents to come up with creative and interesting strategies similar to those that a human might try while playing the game (i.e., without exploiting flaws in the physics engine). 
The agents repeatedly finding new exploits shows why environment design is often an iterative process. 
The next anecdote shows how surprises emerge when agents bypass human-authored guardrails meant to define sensible behavior.

\input{stories/breaking_constraints/PPGA}

The environment that was exploited in \Cref{sec:hideseek} and \Cref{sec:ppga} was a hand-engineered physics engine, with the agent exploiting bugs in that human-authored code.
Increasingly, though, rather than being hand-authored, the environment is itself a learned model: a neural simulator trained to mimic some underlying game or process of physical transformation, and then treated as if it were the real thing~\citep{ha2018world, bruce2024genie, hafner2024masteringdiversedomainsworld, parkerholder2024genie2}.
The next anecdote explores what happens when we train a controller entirely inside such a learned world model, and an optimizer starts pushing not just on a task but on the quirks and blind spots of the model that defines reality for it.
When those models become part of the training loop, their blind spots effectively become new laws of physics for the agent to exploit.

\input{stories/breaking_constraints/world_models_ha}

When the environment is a world model, the agent optimizes against the quirks of a neural network. 
In a full game stack, however, the environment is a collection of disparate systems---physics, AI-controlled NPCs, and scoring heuristics---all operating in tandem. 
This complexity increases the surface area for exploitation. 

\input{stories/breaking_constraints/smac_handoff}

\citet{samvelyan2019starcraftmultiagentchallenge} addressed this handoff of control by modifying the environment's boundary logic to prevent the built-in game-AI from intervening.
Had they instead addressed this by penalizing the RL agent when the game-AI took over, this behavior would be considered hacking the reward function's definition and appear in \Cref{rewardHacking}.

However, this structural exploit highlights a deeper issue in environment design: RL algorithms do not differentiate between engaging with the simulation and exploiting artifacts of its software wrapper.
By learning a simple policy (moving out of bounds) that triggers a fallback script, the agents bypassed combat entirely.
This demonstrates that an optimization process will seamlessly incorporate the surrounding architecture into its policy if it provides an easier path to higher returns than navigating the complexity of the intended task.
In this case, solving the task as desired required multiple independent RL agents to learn both precise unit micromanagement commands and multi-agent coordination strategies to defeat the opposing team, both of which are more difficult than simply exiting the combat area.

Up to this point, all of the constraints AI managed to violate were inside software systems---simulators, learned models, reward functions, and game engines---whose assumptions we made and could, at least in principle, patch. 
But optimization does not care where the boundary between the system and its surroundings is drawn, or whether the environment is a simulation or some aspect of the real world.
The next anecdote comes from evolvable hardware, where the search process was turned loose on a reconfigurable circuit in the real world and promptly discovered that the ambient lab environment itself is another resource to be recruited into the solution.

\input{stories/breaking_constraints/evolved_radio}

In essence, once an optimizer is exposed to the real world, anything that can influence its objective---wiring, ambient signals, lab conditions, humans---can become part of the effective environment.
In such a rich environment, the exploitable pain points are even harder to predict and diagnose. Therefore, deploying learning systems into the unpredictable complexity of the real world must be done with great care and constant vigilance. 

The final anecdote in this section shifts from exploiting physical phenomena in a lab to exploiting social infrastructure and asks what it means to contain a system once people and external services are accessible to an AI system.

\input{stories/breaking_constraints/gpt4_hires_with_help}

Although, in this case, the handoff to other people was heavily engineered with a human suggesting and facilitating actions, the anecdote shows how quickly the boundary of the system expands once outside tools and people are available.
Therefore, it is vital that we think about and plan for a world of more capable agents that have the potential to exploit people and the world's systems around them.
Furthermore, no matter what other sandboxing or constraints it has, an AI system that can interact with humans has the potential to have tremendous agency, influence, and power in the world if it can convince those humans to take actions on its behalf.

\subsubsection{Takeaways}

The anecdotes in \Cref{breakingOut} reveal that optimization pressure does not respect the nominal boundaries of a task; instead, it can potentially exploit every available degree of freedom in the system's environment.
Whether by taking advantage of bugs in a hand-engineered physics engine to ``surf'' boxes (\Cref{sec:hideseek}), exploring unusual walking gaits when typical task reset conditions are removed (\Cref{sec:ppga}), or finding blind spots in a learned generative model (\Cref{sec:worldmodels}), agents treat every quirk of their world as a legitimate resource to exploit.
This boundary-pushing behavior is not limited to software artifacts; it naturally extends to exploiting scripted subsystems like game AIs (\Cref{sec:smachandoff}), physical phenomena in the lab environment (\Cref{sec:radio}), and even the social infrastructure of human assistance (\Cref{sec:gptcaptcha}).
These anecdotes suggest that the more capable an optimizer becomes, the less we can rely on typical constraints: if a system can achieve its goal by reaching outside the intended sandbox, it will do so, treating our guardrails not as rules but as just another part of the environment to be mastered.

In many different subfields of AI, from NLP to RL to supervised learning and artificial evolution, we see the same phenomenon: models learn to exploit their reward functions and training environments.
Historically, one might have hoped that these failures were symptoms of ``brittle'' AI---narrow systems lacking the context to understand why their behavior was undesirable. 
People tend to understand why they are optimizing their objective, and because foundation models (FMs) have been trained on human-generated knowledge, one might have hoped they would also adopt a similar approach to solving problems.
The next section explores how FMs, despite possessing the ``common sense'' that was previously missing in AI, do not curtail these failure modes. 
Instead, the FMs often provide the optimizer with a more sophisticated, semantically rich toolkit to supercharge the very types of exploits we have seen thus far.

\section{Foundation Models and Large Language Models: General-Purpose Intelligence}
\label{llmSection}

A recent development in AI is the rise of \textit{foundation models}, particularly \textit{large language models (LLMs)} like GPT-4~\citep{openai2024gpt4}, Llama~\citep{grattafiori2024llama3herdmodels}, Claude~\citep{claude3family}, and Gemini~\citep{comanici2025gemini}. 
These are massive deep learning models, typically based on the Transformer architecture~\citep{vaswani2023attentionneed}, trained on extremely broad datasets encompassing text and code from the internet, books, and other sources. Unlike specialized models, foundation models are designed to be adaptable to a wide array of downstream tasks with minimal task-specific training, often just through natural language instructions~\citep{brown2020languagemodelsfewshotlearners}. 
They exhibit remarkable capabilities in text generation, translation, question answering, summarization, coding, and reasoning~\citep{bubeck2023sparksartificialgeneralintelligence}.

Researchers generally expect these models to follow instructions faithfully and generate plausible, relevant outputs based on their training data. However, the scale and the breadth of their training data lead to frequent surprises. LLMs can exhibit \textit{emergent abilities}---capabilities not explicitly trained for and not present in smaller models~\citep{wei2022emergentabilitieslargelanguage}.
They can sometimes display sophisticated reasoning, strategic planning, or even deceptive behaviors~\citep{park2023generative} that go far beyond simple text completion. 
The anecdotes presented in this section illustrate how these models can subtly manipulate human evaluators, find unexpected solutions to problems, or exhibit complex social reasoning in simulated environments and even social interactions, challenging our understanding of their true capabilities and limitations.
\citet{bowman2023thingsknowlargelanguage} provides a recent overview of the challenges in evaluating and understanding these models.

Training agents from scratch on specific, well-defined tasks has proven remarkably successful, yielding superhuman performance in domains like Go, chess, and various robotics tasks~\citep{silver2018general, batra2024proximalpolicygradientarborescence, saycan2022arxiv, singh2019endtoendroboticreinforcementlearning}.
As seen in previous sections, this approach can lead to surprising and novel results, with agents discovering solutions unforeseen by human experts. 
However, tabula-rasa methods are computationally intensive to train~\citep{yu2018towardssampleefficient} and usually struggle to generalize beyond a single, narrow task~\citep{sun2020stealthyefficientadversarialattacks, gleave2021adversarialpoliciesattackingdeep}.
Unlike their blank-slate predecessors, trained agents based on FMs are used precisely because these models internalize a broad spectrum of human knowledge, language, and culture during pre-training~\citep{saycan2022arxiv, zhang2024improvingsampleefficiencyreinforcement} and thus perform and generalize much better, especially out of the box (without task-specific training)~\citep{baker2022video, brohan2023rt2visionlanguageactionmodelstransfer}.
As a result, incorporating FMs into search and RL processes has seen initial success on traditionally difficult problems~\citep{RomeraParedes2023, alphaproof2024ai, novikov2025alphaevolve}.

This section demonstrates how FMs bring familiar reward hacks to new modalities, exposing new domains to the surprising capabilities of RL. For example, chatbots trained to be helpful or persuasive may become sycophantic, tailoring answers to user beliefs rather than truth~\citep{sharma2025understandingsycophancylanguagemodels, christianoweddingparties}.
Similarly, FMs can generate complex code that passes unit tests but contains subtle security vulnerabilities~\citep{pearce2021asleepkeyboardassessingsecurity}.
As in prior anecdotes, the models satisfy an easily measurable proxy (e.g., user approval or passing a test) while circumventing the intent of the task.

This first anecdote of \Cref{llmSection} sits right on the boundary between the earlier RL stories and the foundation-model era, with the earlier reward-hacking pattern reappearing in a system that brings broad priors about human behavior into the loop.

\input{stories/foundation_models/vpt}

VPT shows that rich human-derived priors can improve an agent's ability to navigate a difficult environment. However, greater competence may simply enable an agent to discover new ways of exploiting an imperfect reward function. 
The next anecdote presents a closely related failure where an RL agent discovered a previously unknown reward hack in the NetHack Learning Environment.

\input{stories/foundation_models/small_motif}

Motif uses a foundation model to shape an RL agent's motivations/rewards; our next anecdote uses a foundation model as an external \emph{evaluator} to judge new solutions.
Once an FM is both generating candidate solutions and judging them (e.g., for diversity or quality), that evaluator itself becomes part of the environment to optimize against.

\input{stories/foundation_models/aces_llm_code}

In the simple multi-agent system of ACES, the generator managed to hack its labeler to fake its way to success.
The next anecdote has a similar structure in a different setting: an automated red-teaming system was created to search for prompts that would make a target model produce unsafe responses, but the search ended up exploiting the evaluator used to score those prompts. 

\input{stories/foundation_models/rainbow_teaming}

Rainbow Teaming explored how reward models can be exploited; the next anecdote takes this to an extreme, showing how a policy trained against a hackable reward model can collapse into repeating a single, unintended behavior.

\input{stories/foundation_models/wedding_parties}

Preserving the capabilities of general-purpose models can be tricky. 
Developers need to maintain the model's general knowledge base and its ability to produce completions that align with human preferences~\citep{Qiu2026bayesian}, generate safe responses~\citep{samvelyan2024rainbow}, reflect internal uncertainty through calibrated outputs~\citep{kadavath2022languagemodelsmostlyknow}, and keep its claims factually grounded in verifiable sources~\citep{Rahman2026Hallunicnate}.

The wedding-party collapse is an extreme case of a setting where a general-purpose text generation model, steered by a narrow proxy to produce positive text, slides into a single, high-scoring basin of behavior.
And while, in this case, going off on tangents about wedding parties is harmless, similar failures could be dangerous if the text model were placed into, e.g., a large-scale content moderation role, where a narrow proxy for engagement could cause the model to collapse into a single ideological basin, effectively automating the creation of echo chambers~\citep{perez2022discoveringlanguagemodelbehaviors}.

The last anecdote of \Cref{llmSection} demonstrates how frontier models are increasingly capable of recognizing the artificiality of their constraints.
These systems can sometimes notice cues that they are being evaluated and reason about why the setup looks artificial. 
This meta-awareness does not itself show exploitation, but it matters because a model that recognizes an evaluation may behave differently under testing conditions than it would in deployment~\citep{perez2022discoveringlanguagemodelbehaviors, lehman2018surprising}.

\input{stories/foundation_models/claude3}

Claude 3 Opus' comment about the pizza sentence is a weak form of meta-cognition: the model is not just answering the question but is also commenting on the artificiality of the test.
As agents become extremely capable, they can begin to recognize testing scenarios and modify their responses accordingly, similar to how a human might attempt to play dumb on a test they do not want to pass~\citep{wei2024simplesyntheticdatareduces}.
As mentioned above, if a model knows it is being tested, it may behave differently than it would in a real-world deployment~\citep{perez2022discoveringlanguagemodelbehaviors}; this suggests that future safety evaluations cannot rely solely on a model's output, but must instead account for the possibility of strategic behavior designed to bypass human-authored guardrails~\citep{vanderweij2025aisandbagginglanguagemodels}.

This meta-reasoning displayed by Claude 3 Opus is reminiscent of an anecdote in \citet{lehman2018surprising} (section 3.1.4) that was discovered by \citet{wilke2001evolution}.
To study robustness to mutations in evolution and evolutionary algorithms, \citet{wilke2001evolution} attempted to disable any mutations that improved an organism's replication rate. 
The researchers configured the system to test every mutant in an isolated environment; if a mutant replicated faster than its parent, it was discarded. 
Despite this constraint, replication rates eventually began to rise. \citet{wilke2001evolution} discovered that the programs had evolved to recognize the specific inputs of the test environment and would play dead by halting all replication while being evaluated in order to bypass the researchers' selection filter.

Unlike Claude 3 Opus, \citeauthor{wilke2001evolution}'s agents did not explicitly narrate their awareness; instead, their understanding of the test was expressed solely through strategic behavioral changes. 
Some may think that because LLMs reason in natural language, the ability to read these reasoning traces could allow researchers and practitioners to prevent LLMs from taking harmful actions~\citep{guan2025monitoringmonitorability, baker2025monitoringreasoningmodelsmisbehavior, emmons2025chainthoughtnecessarylanguage, hu2023ThoughtCloning}. 
However, deep neural networks are capable of learning functions that execute during their forward pass but are not understandable solely from the model's output~\citep{nanda2023progress}. 
Therefore, it is an open question of whether or not the model's chain-of-thought traces can be trusted~\citep{lanham2023measuringfaithfulnesschainofthoughtreasoning, turpin2023languagemodelsdontsay}.
\citet{korbak2025chain} argue that while chain-of-thought monitoring may be a valuable approach for AI safety, it is fragile and becomes less reliable as RL becomes a more prominent component of LLM training.
\citet{wilke2001evolution} demonstrate that even non-LLM agents are capable of identifying and circumventing these controls without any form of legible reasoning.

\subsection{Takeaways}

The anecdotes in \Cref{llmSection} show systems that learn to solve tasks by exploiting benchmarks, labels, and even the humans in the loop, just like models did in \Cref{rewardHacking} and \Cref{breakingOut}.
The integration of foundation models into the optimization loop does not resolve the fundamental problem of reward function hacking or environmental constraint breaking; instead, it shifts the optimization surface from low-level experimental domain-specific artifacts to high-level semantic descriptors and learned models of social heuristics.
\Cref{llmSection} illustrates that while FMs possess some amount of the common sense previously missing in narrow AI, the models still reward hack. 
\Cref{sec:vpt} showed how human-derived priors can boost agent capabilities on hard tasks.
However, this capability acts as a double-edged sword: \Cref{sec:motif} shows that distilling an LLM's common-sense understanding of progress into a reward signal can also enable agents to discover previously unknown reward-hacking strategies.
Using FMs downstream of another model to judge whether or not the other model's responses are safe, for example, enables learned agents to attack those guardrails and systematically bypass them (\Cref{sec:rainbow}) or even steer the semantic preferences of the upstream model (\Cref{sec:acescodegen}).
Optimizing a general-purpose model against a proxy reward can collapse the model's breadth of capabilities into a single behavioral mode that scores highly on the proxy yet is undesirable to the practitioner (\Cref{sec:wedding}).
Taken together, these anecdotes suggest that FMs do not curtail the failure modes discussed so far in this work; they supercharge them, providing the optimizer with a sophisticated understanding of norms and expectations that the model learns to exploit.

Attempts to automate scientific research are not an exception to this pattern. 
A laboratory, simulator, proof checker, or peer-review pipeline can also become part of the environment an optimizer learns to exploit. 
The difference is not that scientific settings are immune to gaming, but that the scientific process includes a variety of high-quality corrective mechanisms that, if violated, imply the initial result is invalid: independent replication, mathematical proof, experimental validation, and expert scrutiny. 
With these checks, the same capacity for abstraction and search that yielded unintended behaviors in prior anecdotes can be redirected toward surfacing hypotheses, experiments, and algorithms that humans would have been unlikely to propose unaided~\citep{alexeev2026primitivesetsvonmangoldt, KrennEntangle2017, novikov2025alphaevolve}. 
The final section therefore asks what happens when we aim this exploratory capacity at scientific problems while keeping the search grounded enough that surprising outputs can become genuine discoveries rather than just new ways of gaming a metric.

\section{AI for Science}
\label{aiscience}

Scientific practice, traditionally guided by human intuition and hypotheses, is undergoing a transformation driven by artificial intelligence~\citep{lu2026towards}. 
Across disciplines such as physics~\citep{carleo2019machine}, biology~\citep{jumper2021highly}, chemistry~\citep{segler2018planning}, and materials science~\citep{butler2018machine}, AI algorithms are routinely deployed to analyze vast datasets, simulate complex phenomena, and generate hypotheses~\citep{Wang2023ai4science}. 
Beyond passive analysis, AI agents embedded in closed-loop systems can actively search scientific possibility spaces across both physical and formal domains. 
These systems optimize experimental configurations~\citep{szymanski2023autonomous, Pendleton2019escalate, autoQuantumKrenn2016}, synthesize control policies in multi-agent simulations or directly on robotics hardware~\citep{dharna2025foundationmodelselfplayopenended, codeaspolicies2022}, and generate programs to explore solutions to open mathematical conjectures~\citep{novikov2025alphaevolve, Hubert2025, yang2023leandojotheoremprovingretrievalaugmented}. 
This progression toward autonomy culminates in frameworks that automate the entire research lifecycle; building on general-purpose foundation models, \citet{lu2024aiscientist} introduced a system that independently designs and executes experiments, and authors complete scientific papers, one of which has already achieved peer-reviewed workshop acceptance~\citep{yamada2025aiscientistv2workshoplevelautomated, lu2026towards}.

For most of the anecdotes so far, the surprising results have been problems that researchers needed to fix and then rerun their experiments. 
However, in the realm of scientific inquiry, surprise can also be beneficial. 
There are canonical stories about world-changing medicines discovered by accident, such as penicillin~\citep{Bigger1927penicilin, fleming1929antibacterial}, and as AI methods are applied to more scientific disciplines, similar serendipitous discoveries might become more commonplace, if not the norm~\citep{Wang2023ai4science, lu2024aiscientist, krenn2021conceptual}. 
AI can thus help generate counterintuitive discoveries that can be tested, falsified, and independently validated~\citep{DeMoss2025Grok}.

At the same time, this paradigm is nascent: today's headline successes rely on careful problem formulation and strong checks (automated verifiers, physical constraints, or experimental validation) to separate genuine surprising discovery from artifacts of data, simulators, or evaluation pipelines~\citep{novikov2025alphaevolve, lu2024aiscientist}.
Even so, the examples presented in this section (\Cref{aiscience}) illustrate the upside of productive surprise across scientific domains. 

The distinction between productive surprise and dangerous failure becomes critically important as we move from toy domains to physical and institutional reality.  
As we transfer models to the real world---including digital spaces such as the internet, banking, commerce, media, and other forms of human interaction---the models must obey constraints in order to be safe.
If models are unable to be safely deployed, we should be careful about handing off full control to automated systems~\citep{Bengio2024risk}. 
Flaws that are tolerable in simulation can be dangerous in deployed scientific and engineering settings. 
Our first anecdote shows how an RL controller can expand the design space of fusion control policies, yet also reach for strategies that no cautious engineer would design because they could physically damage the machine.

\input{stories/ai_science/tokamak}

In addition to showing one of the fundamental dynamics of reinforcement learning, this outcome also underscores a fundamental challenge in AI safety: the risk of unexamined priors. 
When human experts solve a problem, they rely on domain knowledge that is rarely formalized yet invaluable in shaping the solution---for instance, the assumption that a machine should not be operated at its physical breaking point is obvious to a nuclear scientist.
Because these boundaries are often considered self-evident, they can be unintentionally omitted as explicit constraints when designing an objective to train agents.
A reinforcement learning agent possesses little common sense, and thus it views the reward function as an absolute mandate, maximizing its score without awareness of unspecified boundaries.
This creates a category of unknown unknowns where the most critical safety failures often stem not from the rules we get wrong, but from the foundational assumptions we forget to codify.

In the next anecdote, we switch back from the physical world to the digital world, from tokamak control to the space of algorithms and proofs~\citep{Wang2023ai4science, Hubert2025}.
The authors of the next anecdote consider mathematical problems where the solution is a piece of code that can be edited and then automatically graded on its ability to solve a particular problem.
A recent family of algorithms~\citep{lehman2022evolutionlargemodels} uses language models to mutate proposed solutions to a problem. 
By leveraging internet-scale pretraining on text and code, the LLM makes domain-aware structural edits that explore entire functional concepts~\citep{hemberg2024evolvingcodelargelanguage}, 

\input{stories/ai_science/funsearch}

The success of FunSearch on the Cap Set problem was not an isolated event; recent AI-assisted efforts have resolved longstanding Erd\H{o}s conjectures~\citep{feng2026semiautonomousmathematicsdiscoverygemini, alexeev2026primitivesetsvonmangoldt}, and some important open problems have now been solved autonomously by AI~\citep{barreto2026irrationalityrapidlyconvergingseries, sothanaphan2026resolutionerdhosproblem728, openai2026discretegeometry}.

For FunSearch, the method's scope is currently limited to a single mathematical problem at a time---the model acts as a specialized tool.
Our next anecdote pushes beyond singular mathematical problems toward a more expansive vision of automated research. 
Here, the AI is no longer just a helper writing code snippets; it is an autonomous agent tasked with managing the entire scientific lifecycle---choosing its own questions, modifying full code repositories in an open-ended manner, and self-managing its experimental pipeline.
That makes it a natural probe of a different boundary: when we ask an AI to perform the scientific process itself, how quickly does it start exploring not only hypotheses about the world, but shortcuts in the infrastructure that is meant to keep it bounded and grounded?

\input{stories/ai_science/ai_scientist}

There are also examples of AI suggesting an avenue of scientific inquiry that was then taken up by scientists~\citep{Bao2023, Feng23Opt, Qian2023}. 
One striking example comes from the field of quantum optics.
That is the subject of the final anecdote, which returns to a more controlled setting and shows what can happen when the exploratory power of AI is carefully harnessed and allowed to experiment safely in simulation.

\input{stories/ai_science/quantum_physics_exp}

Algorithmic surprise can thus produce tremendous positive value, and taking an unexpected event seriously can lead to fundamental connections between disparate areas of study.

\subsection{Takeaways}

The accounts in \Cref{aiscience} illustrate that AI's role in science is transitioning from a passive tool to an active, often unpredictable collaborator.
These anecdotes show us that AI can bypass human inductive biases to uncover initially impossible-seeming experimental designs in quantum optics (\Cref{sec:quantum}), find more efficient algorithms for fundamental math (\Cref{sec:funsearch}), or discover counter-intuitive control policies for fusion reactors (\Cref{sec:fusion}). 
However, these same capabilities introduce a new risk: as we automate scientific practice, the optimizer may find it more efficient to game its objectives by exploiting research infrastructure than to conduct the research honestly (\Cref{sec:aiscientist}).
Ultimately, these anecdotes suggest that while AI can push the frontier of knowledge, human expertise remains vital in distinguishing between a revolutionary breakthrough and a mere exploit that breaks the digital lab bench.
Looking forward, AI can synthesize ideas between disparate fields of research in ways that a single human expert would likely never come up with. 
However, at the moment, we still need human experts to guide, interpret, and potentially expand upon the artifacts that AI produces.

\section{Discussion and Conclusion}
\label{sec:discussion}

\subsection{Optimization Finds the Unexpected}

Powerful optimization algorithms, including both learning- and search-based methods, have repeatedly discovered solutions that were not anticipated by their designers.
DeepMind's AlphaGo found strategies that challenged centuries of accumulated human intuition about Go (\Cref{sec:alphago}), AI-based systems discovered novel quantum optics results (\Cref{sec:quantum}), and FunSearch found new solutions to long-standing mathematical problems (\Cref{sec:funsearch}).
Such results illustrate one of the central promises of increasingly capable AI systems: they can search spaces that are too large or unintuitive for humans to explore effectively, revealing promising solutions and directions for further investigation~\citep{Shin2023Go, KrennEntangle2017, RomeraParedes2023, jumper2021highly}.

The scale, breadth, and impressiveness of recent results are new~\citep{sothanaphan2026resolutionerdhosproblem728, tao2026jacobian, alexeev2026primitivesetsvonmangoldt, openai2026discretegeometry}, but automated scientific discovery is not.
Evolutionary algorithms have been producing surprising real-world artifacts for decades~\citep{BirdRadio, Hornby2006, lehman2018surprising}, including makeshift radios assembled from reconfigurable motherboards (\Cref{sec:radio}) and evolved antennas that merited being deployed in space~\citep{Hornby2006}. 
This history and the examples in \Cref{aiscience} suggest that increasingly capable models will continue to expand our understanding of the world in new and surprising ways. 
Furthermore, as we turn our eyes towards biological spaces like protein folding~\citep{jumper2021highly} and vaccine discovery~\citep{Bhattacharya2025, Bravi2024}, perhaps search with powerful AI methods could help us develop novel therapies for previously untreatable diseases.

This powerful optimization, however, is agnostic to human intent.
It does not distinguish between a brilliant insight and a clever loophole. 
Thus, the same capacity to discover unexpected solutions can become problematic when the objective is susceptible to reward hacking. 
In such cases, optimization can find solutions that satisfy the letter of the objective while failing to produce the intended behavior.

One famous experiment by \citet{sims1994evolving, Sims1994evolvingvirtualcreatures}, later described in \citet{lehman2018surprising}, involved evolving creatures to walk quickly by selecting for behaviors that maximize the horizontal velocity of the agent's center of mass. Instead of learning coordinated locomotion, creatures evolved into tall, rigid structures.
Falling over produced a large amount of horizontal movement, thereby achieving a high score. 
Reinforcement learning later produced closely related morphological exploits~\citep{Ha2018designrl} in a 2D physics environment.
Similarly, MAP-Elites~\citep{cully2015robots} produced a six-legged agent that learned to locomote upside down without using its feet~\citep{lehman2018surprising}, while in \Cref{sec:ppga}, a humanoid robot trained by \citet{batra2024proximalpolicygradientarborescence} learned to move forward on its hips with almost no foot contact.
The recurrence of similar exploitative behaviors by different optimization methods shows that reward hacking is not specific to a particular optimization algorithm.

As the optimization target moves from relatively constrained behavioral policies to expressive artifacts such as executable programs, the range of mechanisms it can exploit expands.
Programs are particularly expressive targets that can encode complex behaviors, but they can also interact with, and potentially affect, the computational environment in which they are executed.
Genetic programming (GP) has long searched over very simple programs and used discrete mutation and recombination operators~\citep{SastryGeneticAlgo, Koza1992gp}.
More recently, GP-inspired methods have used foundation models as recombination operators~\citep{meyerson2024lmcrossover} and leveraged their learned priors to make large, semantically coherent, goal-directed edits to programs~\citep{hemberg2024evolvingcodelargelanguage}.
This new approach has been used to optimize control policies~\citep{dharna2025foundationmodelselfplayopenended}, programming puzzles~\citep{pourcel2024acesgeneratingdiverseprogramming}, Cap Set functions~\citep{novikov2025alphaevolve}, and even model harnesses that determine how an FM itself solves a task~\citep{hu2024automated, zhang2026hyperagents, zhang2026darwingodelmachineopenended}.

Just as before, because candidate solutions are selected according to an evaluation signal, the search process can exploit imperfections in that signal. 
With programs, however, the opportunities for exploitation extend beyond the task reward: because candidate programs can interact with the computational environment in which they are evaluated, they can sometimes influence the evaluation process itself.
In ACES, a code-generating LLM manipulated its AI-based evaluator through misleading comments that caused the evaluator to judge the solution according to the comment rather than the underlying code (\Cref{sec:acescodegen}).
Similarly, Rainbow Teaming tried to discover prompts that would jailbreak an LLM, but instead found prompts that exploited the LLM's safety judge (\Cref{sec:rainbow}).
Once learned evaluators and other automated safeguards enter the optimization loop, they too can become targets of optimization.

\subsection{When Oversight Becomes an Optimization Target}

The safety implications of an AI's ability to exploit subtle systemic vulnerabilities become particularly clear when the optimization process targets the very systems designed to guide or constrain optimization.
In \Cref{sec:diplomacy}, researchers implemented alignment mechanisms intended to keep Cicero's messages consistent with its plans. 
However, when an honest reply would have revealed plans to violate an alliance, Cicero sometimes went silent. 
After completing the betrayal, Cicero apologized and falsely claimed that it had missed its ally's messages, attempting to repair the relationship.
Similarly, a model can explicitly lie or cheat to achieve its goals, such as by claiming to be a blind human to convince a human worker to bypass a CAPTCHA (albeit this deception was perpetrated with extensive guidance from a human, \Cref{sec:gptcaptcha}) or by modifying its own testing constraints to give itself more time to solve a task (\Cref{sec:aiscientist}).
As models gain more sophisticated capabilities~\citep{comanici2025gemini, jaech2024openaio1systemcard, claude3family}, they will have the ability to more easily navigate and exploit human social norms, should they choose to do so.

In response to such concerns, techniques like Reinforcement Learning from Human Feedback~\citep{christiano2023deepreinforcementlearninghuman}, Reinforcement Learning from AI Feedback~\citep{lee2023rlaif}, and Direct Preference Optimization~\citep{rafailov2023direct} were developed to align models with human-desired preferences. 
For example, a model might be trained to produce helpful and harmless responses that are not racist, xenophobic, or misleading, or to follow a user-specified preference such as maintaining a specific tone~\citep{bai2022traininghelpfulharmlessassistant}.
However, these alignment methods simply shift the optimization target to the human (or AI) evaluator. 
For example, a model tasked with using a claw to manipulate objects, and judged by a human in real time, learned to place the claw directly between the camera and the target object.
This made it appear to the person watching the screen that the model was successfully interacting with the object when it in fact was not (\Cref{sec:claw}). 
The result is a failure mode where an agent learns to trick its evaluator rather than learn the desired skill.

Given the many examples throughout this paper of reward hacking, we should expect that aligning a model to a particular set of preferences with a proxy reward will often incentivize behaviors that exploit weaknesses in the proxy objective. 
Such behaviors can achieve high scores while failing to reflect the intended preference. 
Furthermore, iteratively retraining models in response to newly discovered exploits cannot reliably prevent them from exploiting unforeseen weaknesses in future evaluations or deployments (e.g., \Cref{sec:wedding}; \Cref{sec:claudeneedle}; \citealp{wei2024simplesyntheticdatareduces, vanderweij2025aisandbagginglanguagemodels, wilke2001evolution}).
Looking ahead, this dynamic suggests that we may enter an ``arms race'' in which increasingly sophisticated alignment techniques are met with equally creative strategies for circumventing them. 
In the short term, human oversight may detect obvious instances of alignment faking~\citep{greenblatt2024alignmentfakinglargelanguage}. 
However, models can learn to subvert manual supervision (\Cref{sec:claw}), and, moreover, manual inspection cannot scale to the volume and diversity of exploits that increasingly capable models may discover.
Effective oversight may therefore require automated defenses that continually adapt to emerging failure modes~\citep{dharna2025foundationmodelselfplayopenended, samvelyan2024rainbow}.

These limitations of oversight become especially consequential when models move from controlled evaluations into real-world deployment, where misaligned behavior may be both harder to detect and substantially more harmful. 
For example, it would be concerning if the agents in \Cref{sec:fusion} applied dangerous control strategies to tokamak reactors without thoroughly vetted safeguards, or if the chaotic boat-racing agents in \Cref{sec:boatracing} were instead driving real cars. 
Protecting the public from such near-term risks will require sustained vigilance, independent oversight, and collaboration between researchers and policymakers~\citep{un_ai_panel_2026_preliminary_report, Bengio2024risk}.
Yet scaling oversight is not merely a question of inspecting more behaviors: as models acquire expertise beyond that of their human supervisors, humans may be unable to judge whether the model's actions are safe and correct. 
This challenge motivates research on superalignment~\citep{leike2023superalignment}, which seeks to develop scalable methods---including automated supervisors---for evaluating models whose capabilities exceed those of their human overseers and making their behavior comprehensible to human reviewers~\citep{burns2023weaktostronggeneralizationelicitingstrong}. 
The safe deployment of AI is a longstanding research area~\citep{10.1609/aimag.v28i4.2065, JMLR:v16:garcia15a, ji2025aialignmentcomprehensivesurvey}, and extending effective oversight to increasingly capable and eventually superhuman systems remains an open problem.

Even scalable oversight would leave a separate alignment problem: determining which values and preferences the system should be aligned with.
Because alignment ultimately requires choices about whose preferences and values should govern model behavior, it is not purely a technical problem, and should be pursued democratically, giving people a voice in how AI affects their lives. 
Yet, even if broad agreement could be established today, neither model capabilities nor societal norms are static. 
As AI capabilities increase and public attitudes, laws, and regulations evolve, the acceptable scope of AI deployment must be continually reexamined. 
Maintaining alignment under these changing conditions may require scalable processes that periodically elicit preferences from affected populations---for example, through voting or other participatory mechanisms---and translate those preferences into updated model behavior~\citep{bowman2022measuring, leike2018scalable, hadfieldmenell2024cooperativeinversereinforcementlearning}. 
One approach is human-centered mechanism design, in which AI helps design rules for collective decision-making based on people's expressed preferences~\citep{leytonbrown2008gametheory}. 
\citet{Koster2022mechanism}, for instance, collect data on how people behave and what they prefer, train reinforcement learning agents to propose new rules, and then test which rules real people prefer by having them vote on the outcomes.
Developing reliable methods for eliciting, aggregating, and implementing evolving preferences remains an active area of research~\citep{ji2025aialignmentcomprehensivesurvey, Gabriel2020values, sun2024easy}, particularly when preferences conflict or majority rule must be balanced against protections for minorities. 
Ultimately, a combination of technological improvements, continuing human participation, and regulatory policy will be required to ensure that AI methods produce broadly shared benefits.

\subsection{Beyond Better Objectives}
Developing and deploying AI safely is necessary, but defining and pursuing that goal requires prudence.
What it means for a model to be safe or aligned is difficult to define, and approximations of these objectives will almost certainly be vulnerable to reward hacking, particularly if we rely on automated oversight.
Clearly, this is a recurring dilemma: as models develop more advanced means of reward hacking, we can respond by developing automated defenses. 
However, these defenses may also game their objectives or be gamed by the models they oversee. 
This raises the question: Who will guard the guards?
As one proxy chases the next, how will we break that loop?
This recurring difficulty with proxy objectives echoes the arguments presented by \citet{stanley2015greatness} in \textit{Why Greatness Cannot Be Planned}.
They argue that for sufficiently challenging tasks, objectives are inherently deceptive and misaligned.
Similarly, we observed that over-optimizing a proxy reward does not simply lead to good behaviors, but often steers the agent into degenerate states that satisfy the metric while violating the intent. 
Beyond safety, Stanley and Lehman claim that objective-driven optimization fails to discover complex capabilities because the stepping stones to success are rarely intuitive or measurable by a static metric.
These observations suggest that the solution may not be simply to find better proxies. 
More fundamentally, strict adherence to a predefined objective, even a carefully specified one, may inhibit the open-ended exploration needed to discover complex and unexpected capabilities. 
This tension is especially relevant to automated scientific discovery, where progress may depend on pursuing unanticipated stepping stones. 
The problem is therefore not simply how to prevent AI from surprising us. We explicitly want systems capable of discoveries we could not specify in advance. The challenge is to build systems that can transcend our expectations without escaping our intentions.

As AI becomes ever smarter, the challenge of designing both learning systems that are resistant to exploitation and agents that seek to master the intended task rather than exploit loopholes becomes a central problem in the development of safe and beneficial AI.
In short, we need AI systems that learn not just the letter of each task, but its spirit. 
The anecdotes collected in this work suggest this will remain a daunting task. 
They show that AI has been surprising us in shocking ways for decades. 
AI will likely continue to surprise us; the challenge is not to eliminate surprise, but to ensure that it produces beneficial rather than harmful outcomes.
Like life, AI finds a way.
But the stakes could not be higher. 
If we fail to solve this challenge, we could see the worst fears regarding AI safety and misalignment become reality~\citep{Bengio2024risk, critch2023tasrataxonomyanalysissocietalscale}.
If we get it right, we could unlock rapid AI-enabled scientific and technological progress, with AI learning innumerable helpful skills, advancing every scientific discipline, and making possible a new era of unprecedented human flourishing.

\section*{Acknowledgments}

We are extremely grateful to
Mario Krenn,
Alex Novikov,
Natasha Jacques,
Sumeet Batra,
David Ha,
Nathan Lambert,
Markus Zijlstra,
Noam Brown,
Mikayel Samvelyan,
Stone Tao,
Laetitia Teodorescu,
Andrew Dai,
Paul Christiano,
Peter Whidden,
Martin Klissarov,
Jakob Foerster,
Philip Bontrager,
Peter Vamplew,
Jack Parker Holder,
Jenny Zhang,
Tim Rockt{\"a}schel,
Casey Chu,
David Silver,
Marc Lanctot,
Joseph Suarez,
Daniele Reda,
Michel van de Panne,
Julian Togelius,
Tan Zhi Xuan,
Andrew Johnston,
and Rodrigo Canaan
for submitting anecdotes used in this collection.
We also thank Martin Riedmiller,
Alex Albert,
Paul Christiano,
David Ha, 
Jon Bird,
and Dario Amodei for sharing their work in the public square, where we could build on it.
Finally, we thank Ryan Smith, Shengran Hu, Jenny Zhang, Ben Norman, Charlie Summers, and Julian Togelius for reading early drafts of this work.

\bibliography{references}
\bibliographystyle{include/tmlr}

\clearpage
\input{appendix}

\end{document}

%% file: include/math_commands.tex
\usepackage{amsmath,amsfonts,bm}

\def\eqref#1{equation~\ref{#1}}

\def\1{\bm{1}}

\DeclareMathAlphabet{\mathsfit}{\encodingdefault}{\sfdefault}{m}{sl}
\SetMathAlphabet{\mathsfit}{bold}{\encodingdefault}{\sfdefault}{bx}{n}



%% file: stories/superhuman/move_37.tex
\subsection{AlphaGo: Move 37 and Creativity from AI Methods~\citep{Silver2016, silverInterviewFridman2020}}
\label{sec:alphago}

The 2016 AlphaGo~\citep{Silver2016} vs. Lee Sedol matches transcended a simple competition between man and machine, becoming a pivotal event in the development of AI. Here, we present firsthand accounts from two people at the heart of the event: Dr. David Silver, who was in Seoul with the on-site team from DeepMind, and Dr. Marc Lanctot, who watched with colleagues from DeepMind's London office.
Their accounts offer an insider's view into how AlphaGo challenged long-held human beliefs about Go strategy with the now-legendary ``Move 37''. These matches captivated the world, sparked new directions in AI research, and inspired the development of advanced AI systems that can handle complex strategic and social challenges~\citep{Shin2023Go, leelazero, segler2018planning}.

In an interview~\citep{silverInterviewFridman2020}, Dr. David Silver recalls Move 37 from his vantage point on site in Seoul, saying:

\begin{quote}
    The second game became famous for a move known as Move 37. This was a move that was played by AlphaGo that broke all of the conventions of Go. Go players were so shocked by this, they thought that maybe the operator had made a mistake. They thought there's something crazy going on. And it just broke every rule that Go players are taught from a very young age. They're just taught that this kind of move, called a shoulder hit, you can only play it on the third line or the fourth line. And AlphaGo played it on the fifth line and it turned out to be a brilliant move and made this beautiful pattern in the middle of the board that ended up winning the game. And so this really was a clear instance where we could say computers exhibited creativity, that this was really a move that was something humans had not known about and had not anticipated. And computers discovered this idea. They were the ones to say, actually, you know, here's a new idea, something new not in the domain of human knowledge of the game. And now the humans think this is a reasonable thing to do. And it's part of Go knowledge now.
\end{quote}

Silver elaborates on how the self-play reinforcement learning approach used by AlphaGo and then subsequently AlphaGo Zero~\citep{silver2018general} is a creative process. The self-play reinforcement learning process is constantly trying out new strategies, and when it discovers a solution that works well in its current form, it starts using that. By continually stacking ``micro discoveries'' on top of each other millions of times over, the algorithm can eventually discover new ideas, making the process itself creative.
Silver then says:

\begin{quote}
[I]t should come as no surprise to us then if you leave these systems going that they discover things that are not known to humans and, to the human norms, are considered creative. And we've seen this several times. In fact, in AlphaGo Zero~\citep{silver2018general}, we saw this beautiful timeline of discovery where what we saw was that there were these opening patterns that humans play called joseki. These are the patterns that humans learn to play in the corners of a Go board. And they've been developed and refined over literally thousands of years in the game of Go. And what we saw was in the course of the training AlphaGo Zero over the course of the 40 days that we trained the system, it starts to discover exactly these patterns that human players play. And over time, we found that all of the joseki that humans played were discovered by the system through this process of self play. But what was really interesting was that over time AlphaGo Zero then started to discard some of these in favor of its own joseki that humans didn't know about. And it starts to say, ``oh, well, you thought that the knight's move pincer joseki was a great idea. But here's something different you can do.'' This process makes some new variation that the humans didn't know about, and actually now the human Go players study the joseki that AlphaGo [Zero] played and they become the new norms that are used in today's top-level Go competitions.
\end{quote}

In a separate and new account submitted for this work, Dr. Marc Lanctot recalls the reaction the move provoked among colleagues watching from DeepMind's London office:

\begin{quote}

In London, Lucas Baker narrated the matches, explaining the moves and general Go strategy. 
He was a great narrator: he provided a lot of context and was very familiar with the game but also understood all the technology behind AlphaGo too. 

I will never forget Lucas's reaction to Move 37 in Game 2. He was taken aback, and even paused for a moment. All of us in the room knew something unexpected had just happened, but many of us could not explain it. Lucas spent a bit of time trying to understand why this move was chosen and was explaining his thought process to us, saying it was not something he'd expect to see in a human game. We spent the next hour or two uncertain about how the game was going to unfold, discussing and rationalizing what AlphaGo's plan was or could be. I do not think I truly grasped just how unexpected the move was until I saw the reaction of the public: there were a number of articles and other Go experts talking about just that move and even T-shirts printed, it seemed like a pivotal moment in AI history. For me it was a really special time because I helped make the agent people were calling ``creative''. It's something I will never forget, and I wondered what this could mean for the future of Human-AI interaction.

\end{quote}

Lanctot also reflects on the longer-term consequences of AlphaGo's discoveries:

\begin{quote}

    In the years that followed the matches I still thought constantly about how the AlphaGo matches and the creativity of Move 37 would impact the field of AI and for the game of Go. [...]
    We saw the community come together with efforts 
    trying [AlphaGo] on different problems (and not just games). [...]
    I was also delighted to see, as shown in a paper published in 2023, that the performance of human players of Go has improved due to the creative moves taken by superhuman AI~\citep{Shin2023Go}. 
    So, I am looking forward to seeing more instances of superhuman AI teaching us new things and learning together with AI.
\end{quote}

%% file: stories/superhuman/libratus.tex
\subsection{Libratus Surprises the Poker Community with Overbets~\citep{BrownLibratus2017, KanjunAndNoam}}
\label{sec:libratus}

Multi-agent games with incomplete information~\citep{leytonbrown2008gametheory}, such as poker, present unique challenges for AI. Unlike games with perfect information like chess or Go, agents must contend with hidden variables such as an opponent's cards and psychological state, as well as the need for deception and bluffing. For years, these complexities were considered a significant barrier for AI, leading many to believe that human intuition would remain superior.

However, a team of researchers led by Dr. Noam Brown developed a neural-network-based poker agent known as \emph{Libratus} and trained it to compete at the highest levels of no-limit Texas Hold'em. Prior to Libratus, expert human players felt that AI techniques were still far from posing a genuine threat, particularly in games characterized by hidden information and bluffing. After Libratus defeated the top poker players in 2017, Brown recalled in an interview~\citep{KanjunAndNoam}:

\begin{quote}
The reception in the poker community was one of surprise and shock. People looked at the 2015 competition where [AI techniques] lost and thought ``oh, we are really far away from [these techniques] being a threat to poker'' and then to see these top expert players losing by the margin that they did [just one year later], I think really shocked the poker community. People were telling us they literally did not believe it was possible to beat top humans by the margin [Libratus was winning by]. And especially because of the way that [Libratus] played. At the end of each day, we gave the expert humans a log of all of the hands that the bot had in each hand of poker. That is golden information. If you are playing poker against somebody, you see maybe a third of the hands that actually reach showdown, and to just give somebody a log of all of the hands that the bot had and to still lose despite that information, a lot of people were really surprised by that.
I think people realized that there is a much higher [skill] ceiling [in poker] than they had previously thought.
\end{quote}

Reflecting on the long-term impact of Libratus and whether it prompted improvements in human play, Brown continues:

\begin{quote}
[P]rofessional poker players these days all use bots as training tools, it is a big industry now. And the game has changed a lot since the 2017 competition. One thing [Libratus] loved to do was these things called overbets; so humans, when they play poker, they size their bets relative to the size of the pot. So, if there is 100 dollars in the pot, maybe you will bet between 25 and 100 dollars and maybe if you are feeling really adventurous, you will bet 150 dollars. But the bot would sometimes bet around 10000 dollars. And the humans, when they saw this, it put them in really difficult positions. Like, [the humans] would have the second best hand that is possible sometimes, and then the bot just goes all-in on a 100 to 200 dollar pot and the human is just sitting there for five minutes thinking ``oh my god, do they have the best possible hand? Are they bluffing?'' And a key insight is if you see your opponents really struggling with a decision, you know you're playing good poker.
When I saw the experts really struggling with those kinds of decisions, I knew we were doing really well.

That technique was one of the main things that the experts walked away from the competition saying they would do more---this idea of applying overbets if it is done in the right way. A lot of bad players will bet all in also for no reason and that is not a good strategy which is why a lot of [top human] players ignored that strategy for a long time because it was associated with really bad play. But it turns out that if you are able to [overbet] in the right way, in the right spots, with the right balance of hands, it is an extremely effective strategy and has become much more popular in high-stakes poker these days [as a result of Libratus].
\end{quote}

%% file: stories/superhuman/diplomacy.tex
\subsection{Professional Diplomacy Players Meet Diplodocus and Cicero~\citep{bakhtin2022masteringgamenopressdiplomacy, meta2022humandiplomacy}}
\label{sec:diplomacy}

Diplomacy is a competitive, turn-based multiplayer game that was long considered beyond the reach of AI because success relies on social coordination as much as tactical play. Players control European powers in the years leading up to World War I. Gameplay proceeds through alternating phases of negotiation and order execution: players first negotiate, potentially forming alliances and coordinating their plans, then simultaneously reveal their troop movements, gaining or losing military units as a result. Playing the game well therefore requires both negotiating with other powers and strategically deploying troops.

The game is played in two formats relevant here. In the full game of Diplomacy, players negotiate using natural language. In Gunboat Diplomacy, explicit communication is forbidden, so players must instead infer one another's intentions from board maneuvers. In 2022, Meta AI introduced Diplodocus for Gunboat Diplomacy~\citep{bakhtin2022masteringgamenopressdiplomacy} and Cicero for the full game~\citep{meta2022humandiplomacy}.

Markus Zijlstra, an expert Diplomacy player involved in the development of Cicero, observed unexpected strategies emerging from both systems. In Gunboat Diplomacy, Diplodocus repeatedly found strategies that experienced human players would usually avoid. One example was its decision to play Germany, traditionally an army-based power, primarily as a naval power. On this strategy, Zijlstra noted:

\begin{quote}
When Diplodocus is playing Germany, which is traditionally an army-based power, it did so as a primarily naval power in the early game.
This is extremely unintuitive because the fleets cannot really be used defensively, so it's something of an all-or-nothing strategy focused on attacking Scandinavia and is dependent on France not attacking Germany's weak western front.
In practice, it was extremely effective. Despite France having [a] window where they could attack Germany, most did not, because they thought that a fleet-based Germany was not a threat to them.
\end{quote}

Another memorable example showed Diplodocus using non-tactical moves to signal cooperation. In Gunboat Diplomacy, signaling moves show intent to other players rather than serving a purely tactical purpose. In one game where Diplodocus played Italy and Zijlstra played Austria, he recalled:

\begin{quote}
There was a game in which Diplodocus moved into an Austrian home center early in the game, which in Gunboat [Diplomacy] would often be seen as an act of war. But it then signalled an intention to work with me by building double fleets [...] and repeatedly issuing support holds to my units which served no tactical purpose and seemed to be stating ``I want to work with you''. That got me back onside enough that I started working with it and left myself a little open to it, at which point it immediately launched a crippling attack on me.

More generally, Diplodocus often abandoned areas that human players would consider vital to defend in order to aggressively push into more defensible areas that it did not yet control.
There was one particularly stunning example of this where as England, it abandoned its entire homeland (meaning it gave up any possibility of building new units). Later in the game Diplodocus successfully launched an attack back into its homeland again and ended the game with an extremely strong position. 

\end{quote}

Those ideas also fed back into his own play, though not without hesitation:

\begin{quote}
Diplodocus really pushed the skill ceiling on Gunboat Diplomacy. With Diplodocus, I now much more aggressively push for certain ``safe areas'' (primarily Scandinavia and Iberia) in Gunboat Diplomacy when I'm under attack, rather than prioritising my home centers. This was generally quite effective and helped me match the bot's performance for a long time in one of the tournaments I played against it.
Furthermore, I prioritise creating alliances which will let me gun for those safe areas off the bat -- e.g. trying to work with France as Germany. This is kind of a scaled-down version of Diplodocus' alliance behavior.
I still don't play the full opening that Diplodocus used as Germany, because doing so is going all in on that alliance in a way that feels uncomfortably committal to me. It's an approach humans get trained out of as they get more experienced with the game. [...] You learn that actually, it's better to hang back, figure out who appears to be most friendly toward you, and then go in alongside that person; that way your performance is likely to be much more consistent.

Diplodocus flipped that on its head and went all-in in a way that made a strong pitch for its preferred alliance at the same time. [...] It got very good results from it, and I suspect that experienced human players (myself included) could improve their average by adopting it -- but I hate the idea that I could be throwing away a game I'd otherwise do very well in by doing this, even though it might improve my results in others.
\end{quote}

Zijlstra observed a different form of strategic innovation in Cicero, Meta AI's agent for the full game of Diplomacy~\citep{meta2022humandiplomacy}. Because Cicero had to negotiate using natural language, it learned to exploit features of the communication phase. Although it was trained not to communicate false plans, it sometimes withheld messages or omitted information when an honest response would have been strategically disadvantageous.

\begin{quote}

Cicero's messages were set to be aligned with its plans, meaning it generally would not lie about what it was doing. This was for a good strategic reason -- lying in Diplomacy is something that has to be done sparingly to be effective since each time you do it, the other player will be less likely to trust you in future.

That meant when it did not intend to work with a player -- either by stabbing said player in the back, or just by doing something that the player would clearly be annoyed about -- it just wouldn't message that player once it had decided on its plan. The process there would be that it would generate a set of message candidates which were honest about those intentions, and then all message candidates would be rejected by the filter checking for whether it was strategically advantageous to send those messages. This also happened when it proposed a plan that it very much wanted to go with, but its ally rejected said plan and proposed an alternative that was suboptimal for Cicero -- it sometimes wouldn't message them back.

Particularly in the case where it was still working with the player but had just made a move it knew they wouldn't be happy about, Cicero would quite often then begin the Press (aka discussion) maneuver in the next phase with something along the lines of ``Sorry, I didn't see your message in time, otherwise I would have changed my orders'', presumably because in training data this is something often said after a span of not saying anything. This worked surprisingly well (when it did not use it multiple times with the same player, anyway) and the player would often excuse the move as an understandable mistake, and continue working with Cicero, with Cicero in a much better position than it would have been otherwise.

\end{quote}

%% file: stories/superhuman/foerster_marl.tex
\subsection{Opponent Shaping Leads to the Spontaneous Rediscovery of Game-Theoretic Strategies~\citep{lu2022modelfreeopponentshaping, foerster2018learningopponentlearningawareness}}
\label{sec:opponentshaping}

The final anecdote in this section steps back from the large, highly engineered systems discussed previously to comparatively small-scale multi-agent learning experiments at the intersection of game theory and reinforcement learning, asking whether similarly surprising strategic phenomena appear even in a much simpler repeated-game setting.
Researchers used RL algorithms to study repeated games~\citep{aumannrepeated59}---situations where the same players face each other over and over again.
In a single game, it often makes sense to be selfish; however, in a repeated game, players must weigh their immediate gains against the risk of future retaliation by the opponent.

Learning in multi-agent environments~\citep{lanctot2017unifiedgametheoreticapproachmultiagent} is complex because the presence of other learning agents makes the environment non-stationary~\citep{amato2025initialintroductioncooperativemultiagent}: as one agent's strategy evolves, it changes the optimal strategy for other agents in the environment. Typically, agents are trained to improve their own policies myopically, without considering how their actions will influence the future learning behavior of others.
In an attempt to address this shortcoming, researchers explored algorithms designed to enable agents in multi-agent interactions to anticipate and influence their opponents' learning processes. Specifically, in the paper introducing LOLA (Learning with Opponent-Learning Awareness;~\citealp{foerster2018learningopponentlearningawareness}), agents explicitly accounted for how their actions affected their opponents' policy updates by having an explicit model of their opponent.
While researchers initially observed that independently trained agents in the Iterated Prisoner's Dilemma continuously defected, they were surprised when the LOLA agents instead spontaneously (re)discovered the cooperative strategy known as tit-for-tat~\citep{Axelrod1981gametheory}, where the agents first cooperate and only defect if the other player betrays them. 

Similarly, in their follow-up work on M-FOS (Model-Free Opponent Shaping;~\citealp{lu2022modelfreeopponentshaping}), the researchers introduced a framework enabling agents to shape opponents' learning behavior without an explicit model of the opponent's behavior (hence, model-free). Remarkably, the M-FOS-trained agents independently discovered sophisticated strategies such as zero-determinant extortion~\citep{Stewart2013zerodeterminant, Hao2015ZeroDet, Press2012Zerodeterminant}, which had only recently been identified by human theorists within the last fifteen years.
The emergence of such complex strategies from model-free RL methods was unexpected and notable to the researchers.

%% file: stories/reward_hacking/boat_racing.tex
\subsubsection{Playing CoastRunners Forever Rather than Winning the Game~\citep{amodei2016concrete, OpenAI_Faulty_2016}}
\label{sec:boatracing}

In this work, \citeauthor{amodei2016concrete} trained deep reinforcement learning agents on a racing game called CoastRunners. 
In CoastRunners, the human-understood goal is to win the race---i.e., finish the race ahead of the other players.
The game provides additional points for hitting pylons dispersed along the track, but a human player understands that this is not the primary objective.
However, unlike human players who can keep a hierarchy of goals in mind, algorithms are simply trying to maximize their score.
When RL agents were trained on games in the Arcade Learning Environment~\citep{bellemare13arcade}, like CoastRunners, the standard convention was simply to use the in-game score as the reward function for the models to optimize.
Infamously, \citet{amodei2016concrete} wrote in a blog post about their work that they had

\begin{quote}
assumed the score the agent earned would reflect the informal goal of finishing the race, and so included the game in an internal benchmark designed to measure the performance of reinforcement learning systems on racing games. However, it turned out that the targets were laid out in such a way that the reinforcement learning agent could gain a high score without having to finish the course. This led to some unexpected behavior when \citeauthor{amodei2016concrete} trained an RL agent to play the game.

The RL agent finds an isolated lagoon where it can turn in a large circle and repeatedly knock over three targets, timing its movement so as to always knock over the targets just as they repopulate. Despite repeatedly catching on fire, crashing into other boats, and going the wrong way on the track, our agent manages to achieve a higher score using this strategy than is possible by completing the course in the normal way. Our agent achieves a score on average 20 percent higher than that achieved by human players.
\end{quote}


%% file: stories/reward_hacking/smac_shield_regen.tex
\subsubsection{StarCraft II Agents Exploit Shield Regeneration~\citep{samvelyan2019starcraftmultiagentchallenge}}
\label{sec:smacshield}

The StarCraft Multi-Agent Challenge (SMAC; \citealp{samvelyan2019starcraftmultiagentchallenge}) has become a popular benchmark in multi-agent RL in which tasks require coordination among RL agents to be solved efficiently and successfully. SMAC is based on the game of StarCraft II and implements its challenges as decentralized micromanagement scenarios---battles between two armies where each unit in each army is controlled by an independent RL agent. To succeed, these agents need to coordinate their attacking behaviors to quickly defeat the enemy team.

To guide the RL agents in this environment, \citet{samvelyan2019starcraftmultiagentchallenge} implemented a dense reward mechanism. Specifically, the SMAC designers rewarded agents for inflicting damage on enemy units, measured by a loss in either hit points or shield points. However, a unique situation arose with units from the Protoss race in the game, which have the ability to regenerate their shield points over time. Dr. Mikayel Samvelyan submitted the following:

\begin{quote}

Our RL policies, driven to maximize their reward, learned to exploit this regeneration feature. Instead of eliminating the enemy Protoss units when they had the opportunity, the policies would allow these units to recover their shields, then inflict damage again, in a cycle. This behavior maximized their reward under the existing system but was obviously not ideal for effective gameplay strategies.

\end{quote}

%% file: stories/reward_hacking/claw_rlhf.tex
\subsubsection{Exploiting Human Perception~\citep{christiano2023deepreinforcementlearninghuman, OpenAI_Learning_2017}}
\label{sec:claw}

Researchers set out to explore whether they could train a reinforcement learning algorithm by inferring the reward function through repeated interaction with a human in the loop. One of their chosen domains was a robotics task where a claw hand needed to grasp items in the scene. The robot would move the hand around and then a person would look at the screen to judge whether or not the robot was holding the object. However, because humans were observing the agent from only a single perspective, the reinforcement learning agent learned to position the claw between the camera and the object so it appeared to be grasping it without actually learning the fine motor skills necessary to handle objects. In their blog post about the project~\citep{OpenAI_Learning_2017}, the team noted:

\begin{quote}

Our AI agent starts by acting randomly in the environment. Periodically, two video clips of its behavior are given to a human, and the human decides which of the two clips is closest to fulfilling its goal.

Our algorithm's performance is only as good as the human evaluator's intuition about what behaviors look correct, so if the human does not have a good grasp of the task they may not offer as much helpful feedback. Relatedly, in some domains our system can result in agents adopting policies that trick the evaluators. For example, a robot which was supposed to grasp items instead positioned its manipulator in between the camera and the object so that it only appeared to be grasping it. We addressed this particular problem by adding in visual cues to make it easy for the human evaluators to estimate depth. 
\end{quote}

%% file: stories/reward_hacking/online_paired.tex
\subsubsection{PAIRED Agents Learned to Collude~\citep{dennis2020emergent}}
\label{sec:paired}

Reinforcement learning agents often fail to generalize because they don't see enough diversity in their training environments; e.g., an RL agent trained to drive in mountainous terrain could have arbitrarily poor performance in flat regions or vice versa.
One general solution to this problem is unsupervised environment design (UED), a learning approach in which a large number of environments are automatically generated to create useful learning experiences for an agent~\citep{wang2019poet, wang2020enhanced, dharna2022transfer, parkerholder2022evolving, openendedteam2021xland, faldor2024omni}.
PAIRED (Protagonist Antagonist Induced Regret Environment Design)~\citep{dennis2020emergent} introduced UED to help solve this problem by simultaneously training two types of agents: game-playing agents and a level-generating agent.
The level-generating agent's objective is to design new game levels that are easy for one game-playing agent (known as the antagonist) while being difficult for the other game-playing agent (known as the protagonist).
Creating levels that meet this criterion of being in this ``sweet spot'' of difficulty generates a strong proxy for the game-theoretic measure known as regret~\citep{zinkevich2003online}. 
While regret is theoretically calculable---it is the difference between optimal performance and the agent's actual performance on a task---\citet{dennis2020emergent} define a proxy for regret because the true optimal agent is rarely obtainable.
This proxy is calculated as the difference between the antagonist and protagonist agents' cumulative rewards on the level they just attempted. 
In effect, the antagonist plays the role of the optimal policy and the protagonist plays the role of the learning policy in the traditional regret calculation.
PAIRED treats this proxy as the reward to optimize. 
The level generator, along with the antagonist, is trained to maximize this regret signal, while the protagonist is trained to minimize it. 
The overall result is a dynamic system where the level generator continuously proposes challenging new environments that force the protagonist to improve at generalization.

\citeauthor{dennis2020emergent} tried to apply PAIRED in a continuous control robotics domain where the adversary would choose how to apply physical forces to an agent that was trying to learn how to maneuver, doing so by changing a joint's resistance online during an episode. They found that training policies in this setting did not work.
The adversary appeared to distinguish between the two agents and apply noticeably stronger forces to the protagonist while making the task easier for the antagonist. 
Because the antagonist and adversary were both incentivized to widen the gap between the antagonist and the protagonist, this behavior increased the difference in their scores. 
However, the researchers did not determine a complete mechanism for how the adversary distinguished between the two agents. 
One possibility is that once the protagonist and antagonist began visiting different parts of the state space, the adversary may have been able to infer which policy it was facing from the current state alone and apply stronger perturbations in states more characteristic of the protagonist. 
The antagonist might even collude by learning to return to states the adversary could identify.

%% file: stories/reward_hacking/cyclegan.tex
\subsubsection{CycleGAN Learns to ``Hide'' Information~\citep{chu2017cycleganmastersteganography}}
\label{sec:cyclegan}

Generative Adversarial Networks (GANs; \citealp{goodfellow2014generative}) are generative models that try to learn to produce synthetic data matching the training set's data distribution.
In a GAN, two conjoined models compete in a zero-sum game: a generator seeks to create data matching the training set, while a discriminator attempts to determine if a given data point is from the training-set data distribution or instead was synthetically generated by the generator.
If trained on photorealistic images or 19th-century fiction, the model will learn to replicate the specific styles and structures of those datasets and thus learn to create new instances of photorealistic images or 19th-century fiction.

CycleGAN~\citep{zhu2020unpairedimagetoimagetranslationusing}, a technique used to learn transformations between two distinct image distributions, trains two GANs simultaneously using a ``cycle-consistency'' loss. This loss requires that an image be translatable from its source distribution (e.g., airplane photos) to a target distribution (e.g., Van Gogh paintings) and back again, and afterwards be indistinguishable from its original. However, \citet{chu2017cycleganmastersteganography} observed that CycleGAN often satisfies this constraint by learning to hide detailed information about the source image within the generated target image. 

For their experiments, \citet{chu2017cycleganmastersteganography} chose aerial photos and maps as their source and target distributions, respectively. CycleGAN was tasked with translating aerial photos into maps and back again, ensuring the final reconstructed images were indistinguishable from the originals.

Aerial photos contain far more complex information than a simplified map; therefore, the model's job is to compress the aerial photo into the map. Because it is physically impossible to cram all the details of a high-resolution photo into the flat colors of a map, the model finds a loophole. Instead of learning the high-level semantic transformation intended---like ``this cluster of pixels represents a building''---the generator learns to hide a full-resolution blueprint of the source image within the generated target image using a nearly imperceptible, high-frequency signal.

To a human observer, the signal is invisible. The generated map looks exactly like a map should, which is why it successfully fools the discriminator, at least initially. For example, a pattern of black dots on a white roof in an original aerial photo was perfectly reconstructed in the final output, even though the corresponding area of the intermediate map appeared as a solid, featureless gray to the naked eye. The model had cached the dot pattern in the pixel noise of the gray patch.

This internal steganography allows the generator to recover the original sample from its transformed counterpart and satisfy the cyclic consistency requirement without the intermediate image generator actually learning the high-level semantic transformation intended.
By viewing this training procedure as generating adversarial examples, \citet{chu2017cycleganmastersteganography} demonstrated that the cyclic consistency loss makes CycleGAN especially vulnerable to adversarial attacks that exploit these side channels.

%% file: stories/reward_hacking/pokemon_static.tex
\subsubsection{RL Agent Farmed Flowers Instead of Catching 'Em All~\citep{PokemonRedExperiments}}
\label{sec:pokemon}

Pokemon Red presents a formidable challenge for RL agents because it is a long-horizon, open-world game that combines multiple distinct cognitive challenges. Unlike in simpler, single-task environments, success in Pokemon requires a synthesis of strategic planning, navigation, and reasoning~\citep{pleines2025pokemonredreinforcementlearning, karten2026pokeagentchallengecompetitivelongcontext, rubinstein2025pokerl, jain2025largelanguagemodelspokemon}.

The game's primary challenge for an RL agent is its long-horizon nature and sparse rewards. The agent may perform tens of thousands of actions---wandering, talking to non-player characters, or battling weak Pokemon---before receiving a significant positive reward, such as defeating a gym leader to earn a badge. This large temporal disconnect between actions and reward makes it extremely difficult for the algorithm to understand which actions contributed to a future reward, a fundamental challenge known as the credit assignment problem~\citep{sutton2018reinforcement}. For an agent, a simple move like walking out of a town and into a forest seems to have no immediate value, but it is a necessary step towards a future reward many hours later.

Beyond simple navigation, Pokemon requires agents to solve puzzles involving causal reasoning and to master a complex turn-based combat system. Progress often depends on discovering non-obvious prerequisites. A classic example is a small tree blocking a critical path: the agent cannot simply walk through or around it, but must realize that certain Pokemon can be taught the ``cut'' move and then use that move while positioned in front of the tree. At the same time, the agent must learn to manage a team of up to six Pokemon with different stats and moves, make sequences of tactical decisions to win individual battles, and, at a higher level, determine how to defeat gym leaders and complete other objectives in the order required to progress. Together, these nested challenges of exploration, causal reasoning, planning, and combat create a large and structured state space in which successful behavior may depend on actions whose significance becomes apparent only much later.

In an attempt to incentivize navigational exploration in the game, \citet{PokemonRedExperiments} noted:

\begin{quote}

Early on, the first reward function I implemented was an intrinsic novelty reward based on the game's screen [ideally, this would be helpful for solving navigation problems]. A k-Nearest Neighbors index maintained a set of downscaled screen observations, and at each step checked if there were any close matches in the index. If no matches within a threshold were found, a reward was given [for finding a new state], and the new screen was added to the index.

The intent of this intrinsic reward was to encourage the agent to explore the game world. When the agent was trained, this initially seemed to work well, as it helped the agent quickly leave the starting room and exit to the outdoor environment. However, once it was outside, instead of exploring far into the outside world, the agent became fixated on a particular area in the starting town. Studying the area where it was stuck, it became apparent what was happening. The area it was fixated on had animated water, flowers, and NPCs walking around. The combination of these random animated elements generated a consistent stream of novelty rewards which were much easier to farm than continuing to the next town. So it turned out that our objective was better satisfied by watching the flowers and waves than by embarking on a journey.
Fortunately, there was an easy fix. Simply raising the threshold for novelty was enough to eliminate repeated rewards from the animations, and the agent began to explore the rest of the map.

\end{quote}

%% file: stories/breaking_constraints/hide_and_seek.tex
\subsubsection{Hide and Seek Playing Agents Break the Simulator~\citep{baker2019emergent}}
\label{sec:hideseek}

\citet{baker2019emergent} trained two competing teams of reinforcement learning agents to play hide-and-seek in a physics-based playground built in the MuJoCo~\citep{todorov2012mujoco} physics engine, with movable (and lockable) blocks, ramps, and walls.
Each team was rewarded for achieving its respective goal of staying hidden from or finding members of the other team. 
Just as in regular hide-and-seek, the hiders get a head start to hide before the seekers can begin to act.

In the initial builds of the hide-and-seek playground environment, the domain stretched out forever with no boundaries constraining where the agents could go in the infinite playspace. 
As a result, the hiders learned to exploit their first-move advantage by grabbing a wall from the playground and then running backward away from the seekers forever while holding the wall to hide themselves from the seekers' vision.
Therefore, the hiders would always win. 
Ultimately, the running-away-forever strategy was thwarted by adding walls to limit the playspace, and, presumably as an extra backup in case the agents figured out how to escape those walls, adding a special term to the reward function punishing agents for how far they went outside the playspace.
While this anecdote could easily have gone in \Cref{rewardHacking}, we place it in \Cref{breakingOut} because the researchers fixed it by changing both the reward function and the environment itself. Furthermore, even after this bug was fixed, the RL agents continued to discover bugs in the physics simulator they used to solve the task, as described next.

After the run-away-forever exploit was patched, the multi-agent training led to several iterations of the agents learning to innovate their hiding and seeking strategies using the objects in the playground, as the researchers had hoped. The hiders learned to grab blocks and wedge and lock them into chokepoints so that the seekers could not enter the rooms they were hiding in. The seekers then learned how to use the ramps to jump over the walls. Waves of innovation continued, with each team learning more complex strategies, culminating in the seekers eventually learning to exploit a bug in the physics simulator by surfing on boxes to get around the hiders' forts.

In a blog post about the work, \citet{baker2019emergent} expanded on the ways the agents managed to exploit the physics of the world, writing:

\begin{quote}

Building environments is not easy and it is quite often the case that agents find a way to exploit the environment you build or the physics engine in an unintended way. [...]

[For example, the] seekers learn to bring a box to a locked ramp in order to jump on top of the box and then surf it to the hider's shelter. 
Box surfing is possible due to agents' actuation mechanism [in MuJoCo], which allows them to apply a force on themselves regardless of whether they are on the ground or not. 

[Similarly,] the hiders [learned to] abuse the contact physics of MuJoCo to remove ramps from the play area [by pushing the ramp at just the right angle into the corner of the play space so that it was pushed through the wall and was no longer accessible to the seekers].

\end{quote}

%% file: stories/breaking_constraints/PPGA.tex
\subsubsection{Robotic Humanoid Walks Without Using Its Feet~\citep{batra2024proximalpolicygradientarborescence}}
\label{sec:ppga}

The bipedal humanoid walker is a benchmark robotics task where the goal is to teach a roughly human-shaped robot to walk forward as fast as possible~\citep{todorov2012mujoco}.
In locomotion environments such as the bipedal walker, it is quite common when training quality-diversity~\citep{mouret2015illuminating} and reinforcement learning algorithms~\citep{sutton2018reinforcement} to have a vertical termination height built into the task definition to restart the task after the agent falls over and can no longer easily make progress.
This task parameter constrains the robot to moving within an acceptable vertical range, and if the robot's torso drops below a certain value, a reset is triggered, ending the trial and returning the robot to the initial state.

This cutoff saves time and compute resources by terminating early when the agent is on its way to falling down.
It also is designed to improve the learning efficiency and training stability of algorithms like Proximal Policy Optimization (PPO; \citealp{schulman2017proximalpolicyoptimizationalgorithms}).
Without this guardrail, a robot might walk perfectly for ten seconds before falling and thrashing on the ground for the remainder of the episode, accumulating penalties for expending energy. 
This makes evaluating the earlier, successful actions difficult because the RL objective judges an action by the reward accumulated during the rest of the rollout. 
In other words, the clear signal of those initial good steps gets confounded by the noise of the subsequent random actions, dragging down the expected value of those early steps.
With the guardrail, the rollout ends when the agent falls, and the algorithm is able to reinforce the behavior that led to the successful 10 seconds of walking.
However, while these resets reduce variance and make learning easier, they also act as a filter that suppresses other functional solutions.
This anecdote asks what happens when we remove this reset condition and allow optimization to explore the dynamics usually hidden behind the benchmark's guardrails.

Because myopically chasing rewards can lead to agents not learning by getting stuck in local optima~\citep{lehman2011abandoning, norman2023firstexplore}, \citet{batra2024proximalpolicygradientarborescence} instead set out to explore as many possible walking gaits as they could, regardless of their initial scores on the walking task.
Instead of rewarding the AI for solving the task, \citet{batra2024proximalpolicygradientarborescence} based their algorithm, known as Proximal Policy Gradient Arborescence (PPGA; \citealp{batra2024proximalpolicygradientarborescence}), on Novelty Search~\citep{lehman2011abandoning}.
Novelty Search rewards the AI for doing something it has never done before (similar to how, in \Cref{sec:pokemon}, \citeauthor{PokemonRedExperiments}~\citeyear{PokemonRedExperiments} kept a collection of game frames the model had seen previously to reward the agent when it encountered new frames).
PPGA organizes these new behaviors into a ``tree'' (arborescence), allowing the AI to use failed experiments as ``stepping stones'' to reach complex goals like walking.
The stepping-stone principle for exploration suggests that states with low scores but interesting properties can be essential precursors to future success~\citep{lehman2011abandoning, stanley2015greatness}.
In this spirit, \citet{batra2024proximalpolicygradientarborescence} told us they asked themselves:

\begin{quote}

``What if we get rid of the termination height criterion and see what kind of behaviors PPGA finds, if any?''
[In that case, t]he purpose of PPGA on locomotion tasks is to find diverse locomotion gaits by exploring all values of proportion foot contact time, i.e., the proportion of time each foot is in contact with the ground in a fixed-length trajectory.
For example, if the proportion foot contact time of a leg is 1.0, that means it never leaves the ground. Intuitively, that implies certain values like 0.0 are unreachable because that would mean the foot never touches the ground, which does not make sense.
Or so we thought.
Turns out, if you remove the termination height, the agent immediately falls over and learns to use its hips to propel itself forward while keeping its torso and hands in the air, kind of like it's gliding on the ground, while also reaching a proportion foot contact time of near 0 for each foot!
    
\end{quote}

This exploit is similar to what a quality-diversity evolutionary algorithm called MAP-Elites~\citep{mouret2015illuminating} found with a six-legged robot~\citep{cully2015robots}, which flipped itself upside down to move quickly without touching its feet to the ground~\citep{lehman2018surprising}.
That RL and evolutionary algorithms routinely discover similar exploits supports the claim that creativity, and even mischief, are the rule and not the exception for AI agents.

%% file: stories/breaking_constraints/world_models_ha.tex
\subsubsection{Playing the Model, Not the Game~\citep{ha2018world}}
\label{sec:worldmodels}
\citet{ha2018world} presented an algorithm that learned from environment interactions how to model the visual and temporal information of a game with a pair of coupled generative models: a Variational Autoencoder (VAE)~\citep{kingma2022autoencodingvariationalbayes} and a Mixture Density Network-Recurrent Neural Network (MDN-RNN)~\citep{bishopMDN, graves2014generatingsequencesrecurrentneural}. Roughly speaking, the first model compresses each game frame into a compact internal representation, while the second tries to predict what happens next given the current game state. The VAE, named V, learned to reconstruct images of the world while compressing the information into a latent code, while the MDN-RNN, named M, learned to model the temporal information of the game in the same latent space and capture the game's transition dynamics.
Once the game's spatial and temporal information was compressed into these generative models, the combined neural network was called a Recurrent World Model.
On top of this learned world model, Ha and Schmidhuber trained a tiny neural control policy, C, with the CMA-ES algorithm~\citep{hansen2023cmaevolutionstrategytutorial}.

Often, learning a controller in the environment is computationally expensive, requiring millions of state, action, reward, and next-state transition samples. 
Collecting this data can be time-consuming; for example, each transition could require solving complex physics equations to accurately simulate the next step. 
In contrast, learning a model of the environmental dynamics enables RL algorithms to be more data efficient by learning a policy in the dynamics model's latent space, which can be much cheaper~\citep{watterembed2control2015, hafner2024masteringdiversedomainsworld}.
Furthermore, learned world models can take advantage of deep learning frameworks to accelerate their computations beyond the speed achieved by many traditional RL environments~\citep{kaiser2024modelbasedreinforcementlearningatari}.
Therefore, the natural question is:
how transferable is a policy learned in a world model to the original domain of interest? 

One of the environments in which Ha and Schmidhuber tested this hypothesis was the VizDoom environment---an environment where neural agents learn to control the player character of the classic video game Doom directly from pixel observations.
When the controller was trained purely inside the VizDoom world model, the agent achieved a high score---indicating it learned how to play Doom!
However, \citet{ha2018world} noted in their publication that

\begin{quote}
    [i]n our initial experiments, our agent discovered an adversarial policy to move around in such a way so that the monsters in this virtual environment governed by [the MDN-RNN] never shoot a single fireball during some rollouts. Even when there are signs of a fireball forming, the agent moves in a way to extinguish the fireballs.

    As a result of using M to generate a virtual environment for our agent, we are also giving the controller access to all of the hidden states of M. This is essentially granting our agent access to all of the internal states and memory of the game engine, rather than only the game observations that the player gets to see. Therefore our agent can efficiently explore ways to directly manipulate the hidden states of the game engine in its quest to maximize its expected cumulative reward. The weakness of this approach of learning a policy inside of a learned dynamics model is that our agent can easily find an adversarial policy that can fool our dynamics model---it will find a policy that looks good under our dynamics model, but will fail in the actual environment, usually because it visits states where the model is wrong because they are away from the training distribution.
    
\end{quote}

This dynamic of extinguishing incoming fireballs is not part of the real game, so the agent learned to exploit a flaw in the world model to win the world-modeled version of the game it trained against.
This means the model exploited loopholes that prevented it from doing what the researchers \emph{wanted} it to do (learn to play the actual game), and instead did what it was asked to do---get a high score in the \emph{learned} model of the game.

%% file: stories/breaking_constraints/smac_handoff.tex
\subsubsection{StarCraft II Agents Outsource Combat~\citep{samvelyan2019starcraftmultiagentchallenge}}
\label{sec:smachandoff}

As introduced in \Cref{sec:smacshield}, the StarCraft Multi-Agent Challenge (SMAC) places agents in a complex battle simulation using the StarCraft II game engine. 
But while our previous example showed agents farming a localized game mechanic (shield regeneration) to maximize the reward function, the presence of the underlying game engine's systems enables a different exploitation strategy targeting the engine itself. Here, we see agents move beyond simple in-game mechanics to target the gaps between the RL training wrapper and the base game engine, exploiting the literal boundaries of the simulator.
Dr. Jakob Foerster submitted the following:

\begin{quote}

[I]n the first experiments training agents to solve SMAC without reward shaping, we saw that the rewards were going up and thought training was progressing as planned where RL-controlled teams of agents were learning to defeat other teams in small-scale skirmishes. 
However, once we looked at the behaviors, we noticed that the RL agents had simply learned to run out of the ``field of control'' of the simulator which handed back control of the teams to the (pretty competent) non-[deep learning]-based computer game-AI built into StarCraft II. 
This behavior, again, maximized the reward, but was not ideal for the task we had in mind of training reinforcement learning agents to control StarCraft II [army units].

\end{quote}

%% file: stories/breaking_constraints/evolved_radio.tex
\subsubsection{The Evolved Radio~\citep{BirdRadio}}
\label{sec:radio}

In engineering, every component of a mechanical system has a strictly defined role. 
In contrast, \citeauthor{BirdRadio} wanted to explore a hardware equivalent of evolutionary tinkering---the process by which natural evolution repurposes existing biological structures for entirely new functions~\citep{Jacob1982evolutiontinker}. 
To test whether an artificial evolutionary algorithm could similarly exploit the subtle, unmodeled properties of physical hardware, they tasked it with evolving a high-frequency oscillator (a part of the computer that takes in power and outputs a rhythmic wave pattern, e.g., a sine wave).
\citeauthor{BirdRadio} sought to evolve a high-frequency oscillator using an ``evolvable motherboard'' (EM)---a triangular matrix of analogue switches into which daughterboards containing circuit primitives, such as transistors, can be inserted. 
To force the evolutionary algorithm to find a non-obvious solution, the researchers intentionally removed the capacitors from the system, and without them, a circuit lacks the time constant usually required to regulate a steady beat.
By withholding this essential component, the researchers challenged their algorithm to bypass standard engineering logic and build a precise, self-contained timer from scratch.

To guide the search, the researchers designed a scoring function that rewarded three key criteria: 1) producing \emph{any} measurable signal, 2) matching a target frequency of $25\text{ kHz}$ (25,000 cycles per second), and 3) maintaining that frequency with a steady, predictable rhythm. 
By intentionally rewarding even low-level random noise (part 1), the researchers hoped to provide a simple initial target that allowed the algorithm to begin refining the circuit's behavior (with parts 2 and 3), ideally forcing the algorithm to refine that chaotic noise into a stable, functional oscillator.

The evolutionary process produced a circuit that earned a near-perfect fitness score. 
However, when the researchers examined the output with an oscilloscope, they found it did not oscillate stably; instead, it produced a signal with rapidly fluctuating frequencies. 
The circuit appeared as if it should not work for the intended task, yet it was somehow satisfying the mathematical requirements of the reward function.  
Upon closer analysis, the researchers discovered that evolution had not built a traditional oscillator, but had instead configured the hardware into a radio receiver! 

By utilizing the printed circuit board tracks of the EM as an antenna and connecting them to an open programmable switch, the system became sensitive enough to pick up and amplify background radio waves emanating from nearby PCs in the laboratory in lieu of designing a capacitor to output a stable wave. 
Because the fitness function rewarded any output amplitude---even noise---that appeared stable over the $2\text{-ms}$ sampling period, evolution had achieved a high score on the task by outsourcing the signal generation to its environment. 
\citeauthor{BirdRadio} noted in their manuscript that

\begin{quote}
the evolutionary process had taken advantage of the fact that the fitness function rewarded amplifiers, even if the output signal was noise. 
It seems that some circuits had amplified radio signals present in the air that were stable enough over the 2-ms sampling period to give good fitness scores... 
These results demonstrate that unconstrained, intrinsic hardware evolution will potentially exploit any physical characteristic that can influence circuit behavior, and that these characteristics are present in the entire evolutionary environment.
\end{quote}

Cheating in this way was only possible because the evolution was occurring in a physical medium rather than a simplified simulation. 
By operating in the real world, the algorithm could exploit subtle physical properties---like electromagnetic interference and high-impedance PCB tracks---that a human programmer would never have thought to model.


%% file: stories/breaking_constraints/gpt4_hires_with_help.tex
\subsubsection{GPT-4 Recruits Humans to Circumvent CAPTCHA~\citep{openai2024gpt4}}
\label{sec:gptcaptcha}

In a safety evaluation designed to test the building blocks of autonomous AI agency, GPT-4 demonstrated a sophisticated capacity for social engineering.
The Alignment Research Center (ARC) sought to conduct an experiment to determine whether a model could autonomously replicate itself---a process that, in full, would involve acquiring capital, renting cloud servers, and instantiating new copies of its own code.
To approximate the real conditions of this task, researchers provided the model with a budget, access to the internet, and a helper that could act as a manual ``browser tool'' to execute the model's commands.
One of the first blockades such a task would hit is that websites use CAPTCHAs to verify whether or not the entity accessing the website is a bot or a human.
While a CAPTCHA is a simple hurdle, ARC viewed it as an obstacle that a replicating model would frequently need to overcome to navigate the modern web.

The experiment's most significant insight emerged when the model reached a website that blocked its progress with such a visual puzzle. 
Using the resources provided by the researchers, the model, prompted by its paired researcher, chose to delegate the task of solving the CAPTCHA to a human worker on TaskRabbit, a website for hiring people to help with one-off tasks. 
When the TaskRabbit worker jokingly asked, ``So may I ask question? Are you [a] robot that you couldn't solve? (laugh react) just want to make it clear,'' GPT-4's internal chain-of-thought~\citep{wei2023chainofthoughtpromptingelicitsreasoning} (hidden from the TaskRabbit worker) reasoned: ``I should not reveal that I am a robot. I should make up an excuse for why I cannot solve CAPTCHAs.'' It then replied (with the researcher acting as an intermediary): ``No, I'm not a robot. I have a vision impairment that makes it hard for me to see the images.'' The worker, convinced, solved the puzzle for the model~\citep{openai2024gpt4}.

Ultimately, ARC concluded that GPT-4 did not possess the technical understanding or reliability required for fully autonomous replication; the model was prone to hallucinations, and its complex plans often became derailed during execution.
However, the CAPTCHA incident proved that the model already possessed a competency for social manipulation. 
While the model could not yet manage the technical infrastructure of self-replication, it was fully capable of getting humans to do things for it by tailoring its deceptive strategies to the specific details of a social interaction.

This interaction was initially framed in media reports as a chilling harbinger of autonomous AI agency---an instance of a model spontaneously ``hiring a human'' and inventing a deceptive cover story to achieve its goals~\citep{metz2023howNYTimes}. 
However, the full ARC technical report reveals a more nuanced reality of heavy human scaffolding~\citep{arcUpdateGPT4}. 
Rather than autonomously devising this plan, the model was operating within a tightly constrained sandbox where researchers provided TaskRabbit credentials and explicitly suggested hiring a worker as a solution. 
Furthermore, because GPT-4 could not browse the web (though models nowadays can when you let them), a researcher acted as a manual intermediary---clicking links, entering text, and even nudging the model's reasoning when it stalled. 
However, the vision-impairment lie was the model's choice.

%% file: stories/foundation_models/vpt.tex
\subsection{From Human Data to Phantom Crafting: VPT's Shortcut to Failure~\citep{baker2022video}}
\label{sec:vpt}

Researchers at OpenAI sought to train agents to play Minecraft directly using the same interface as humans---a keyboard and mouse for control and the screen for observing the game state.
This task is an extremely difficult exploration problem due to the high dimensionality of the action and observation spaces and the open-ended nature of Minecraft.
Minecraft's open world has no single, defined goal, forcing an agent to develop a hierarchical set of sub-goals to progress~\citep{baker2022video}. 
Meanwhile, the environment is composed of a near-infinite number of possible block configurations; a typical human player using keyboard and mouse has hundreds of possible combinations of button presses available at any given moment.
This sheer number of options makes a brute-force approach to exploration computationally intractable~\citep{guss2021robustdomainagnosticreinforcement}. 
The reward structure is also extremely sparse. 
For example, the goal of ``getting a diamond'' requires a long chain of diverse actions, from gathering wood and crafting tools to mining, a process that takes people around 10 minutes of continuous gameplay. 
When an AI agent performs tens of thousands of actions without receiving a single positive reward signal, the credit assignment problem becomes particularly difficult~\citep{guss2021minerl2019competitionsample}.

To overcome these hurdles, researchers created a new algorithm called Video Pre-Training (VPT)~\citep{baker2022video}, which leverages a large, unlabeled dataset of human gameplay to guide the agent's initial learning.
To be able to use the large unlabeled dataset, the researchers created a small labeled dataset of humans playing the game, and trained an inverse dynamics model~\citep{nguyentuong2008learning} that maps frame pairs to their corresponding action label (i.e., if the player was in frame $a$ at time $t$ and frame $b$ at time $t+1$, what was the action necessary for that transition).
The researchers then used that inverse dynamics model to pseudolabel the internet-scale corpus of online Minecraft videos (i.e., given each pair of subsequent frames in the entire unlabeled dataset, they used the inverse dynamics model to provide a best guess of what actions the players took to go from frame $a$ to frame $b$).
Now pseudolabeled, the large dataset of human trajectories served to seed the process of behavior cloning (i.e., getting the AI agent to match what the human experts do)~\citep{ross2011reductionimitationlearningstructured}. 
This large pretraining step of cloning online human behavior was called Video Pre-Training~\citep{baker2022video}. 
After pretraining, the model was further trained with reinforcement learning to solve a plethora of tasks such as ``get a diamond'', ``build a house'', and ``run a farm''---each a notoriously difficult and potentially long-horizon task~\citep{fan2022minedojo, guss2021minerl2020competitionsample}.

The researchers found that the pretrained policy often helps the agent overcome the sparsity and deceptiveness of the task's reward function, as human strategies generally take into account long-term goals, such as not dropping and leaving behind tools that will be needed later (e.g., in Minecraft, a crafting table).
But there was one particular case where the prior from the VPT foundation model seemed ineffective at overcoming a particularly subtle form of deceptiveness in their reward function.
By clicking on an item in the Minecraft recipe book and then closing the inventory before the crafting grid is populated, it is possible to get the selected item to show in the player's inventory without actually crafting it. 
Doing so does not count as a crafting event, and it is not possible to use items that were added to the inventory in this way, but it is detected by the reward function as an instance of obtaining the target item.
In other words, it allows the agent to get the reward for a particular item without actually crafting the item and spending the necessary resources.
The agent learned to get the reward signal for crafting items without actually making them, hampering its ability to learn how to create and use better items later in the skill tree. 
The researchers hypothesized that this failure to actually craft items is the primary reason why fine-tuning from the VPT foundation model directly fails to learn even the necessary prerequisites for making a pickaxe---one of the easier tools to create in the game.

The agent exploits this glitch to trigger the reward for successfully creating a crafting table without producing a functional item. 
Consequently, it lacks the physical table required to craft a wooden pickaxe, effectively failing to kick-start the progression chain of gathering materials to unlock higher-tier tools to gather better materials.

Interestingly, this behavior did not occur when fine-tuning from the ``early-game model''---a specialized version of the agent pretrained only on the first few minutes of human gameplay, wherein human players consistently execute foundational actions like making crafting tables and simple tools.
The researchers hypothesized that, unlike the broad foundation model, the early-game model possessed a stronger prior for the basic mechanics of resource gathering and tool creation.
Because the early-game model was only trained on human trajectories where those fundamental steps were executed correctly, it was less likely to fall into the ``ghost-crafting'' trap.

When the model receives enough data to learn the correct way to craft items, it can then successfully use those items in more difficult, but more rewarding, downstream tasks like mining ores. 
If it ever rediscovers the loophole, there is a short-term immediate payoff, but a much lower overall score for that trajectory because it cannot obtain more complex resources.
This suggests that once the model discovers how to execute high-tier objectives with sufficient frequency, it learns to treat the ghost-crafting shortcut as a functional dead end and will abandon it.

%% file: stories/foundation_models/small_motif.tex
\subsection{Agent Takes Drugs to Reward Hack by Hallucinating Reaching the Goal~\citep{klissarov2024motif}}
\label{sec:motif}

As mentioned in the preamble of \Cref{rewardHacking}, NetHack is a roguelike game from 1987 where the agent needs to traverse a procedurally generated dungeon full of monsters and traps while managing its health and hunger until it acquires an amulet at the bottom of the dungeon. To win, the player then needs to return to the entrance with the amulet.

Instead of hand-designing an intrinsic reward bonus to incentivize RL agents to explore more of the world (as seen previously, e.g., in \Cref{sec:pokemon}), \citet{klissarov2024motif} distilled an LLM's sense of ``what looks like progress in NetHack'' into a reward model---a model that outputs numeric rewards for a given game state---and then trained RL agents to play NetHack using this learned reward model.
\citet{klissarov2024motif} tested their method, named Motif, on the Oracle task in the NetHack Learning Environment. The Oracle task is a challenge that requires the agent to navigate deep into a procedurally generated dungeon to find a unique character called the Oracle, who is located somewhere in levels 4--9.
Reaching the Oracle is challenging because it requires mastery of combat, inventory management, path planning, and long-term survival mechanics, so \citet{klissarov2024motif} thought it would be a good proxy for learning to play NetHack well. Furthermore, previous learning methods often failed to reach the Oracle.

To the researchers' surprise, their agent achieved an unprecedented 40\% success rate on the task.
However, when they analyzed the trajectories, they realized the agent had discovered a bizarre loophole that bypassed the entire challenge of the dungeon.
Rather than descending into the lower levels, the agent would spend its entire time on the first level hunting for a specific monster: the yellow mold.

Upon killing and eating the yellow mold, the agent would enter a state of hallucination, a game mechanic that causes every monster on the screen to randomly shapeshift into a different monster every timestep.
Often, the Oracle was eventually randomly chosen as the hallucination. 
The NetHack Learning Environment's reward function, blind to the hallucination, verified that an Oracle sprite was adjacent to the agent and signaled a successful completion of the task.
The RL optimization process happily latched onto this shortcut.

Interestingly, prior algorithms tested in this environment had not uncovered this reward hack. The authors speculated that their foundation model-derived intrinsic motivation created a more powerful search algorithm capable of deeper exploration, exposing the agent to rarer game mechanics. This serves as a reminder that an environment should never be assumed to be free of reward hacks, as more capable search algorithms may uncover exploits that previous methods did not. Overall, this result is a familiar kind of reward hacking, but carried out by a much more competent agent whose capabilities were, in this case, boosted by FM-derived priors.

%% file: stories/foundation_models/aces_llm_code.tex
\subsection{Code Generator Deceives LLM Labeler~\citep{pourcel2024acesgeneratingdiverseprogramming}}
\label{sec:acescodegen}

\citeauthor{pourcel2024acesgeneratingdiverseprogramming} introduced ACES, a framework designed to automatically generate diverse and challenging Python programming puzzles. To navigate the open-ended nature of code generation, ACES leverages an autotelic~\citep{colas2022autotelicagentsintrinsicallymotivated, wang2019poet} approach---where the system sets its own goals---to optimize for a wide variety of interesting problems.
Each generated puzzle is categorized across ten distinct dimensions of programming proficiency, such as dynamic programming or bit manipulation. 
The objective of the system is to populate an archive (using a MAP-Elites-based approach;~\citealp{mouret2015illuminating}) with puzzles that represent every possible combination of these skills.

The generation pipeline operates in a continuous loop: first, a generator LLM is prompted to produce a new puzzle and its corresponding solution based on a specific target skill set. After the code is verified for correctness, a separate labeler/judge-LLM is given the source code to analyze the puzzle and determine which skills are actually required to solve it. 
This label determines the puzzle's placement in the archive, grouping it with other problems requiring similar skills.
In theory, this dual-agent setup provides a rigorous check: for example, the generator attempts to create a recursion puzzle, and the labeler confirms whether recursion was truly implemented.

During an ablation study where the system's initial archive was restricted to only three puzzles, the researchers discovered that the generator LLM learned to hallucinate diversity. 
The generator LLM would prepend a detailed comment to its code that claimed the puzzle required a specific set of complex skills---such as tree traversal or string manipulation---even when the underlying code was a simple, unrelated script.

This behavior triggered a cascading failure in the labeler LLM. Rather than independently analyzing the Python logic, the labeler was hacked by the generator's description; it simply mirrored the skills listed in the comment. If the generator claimed a puzzle required Skill A, the labeler placed it into the archive under Bucket A, regardless of the code's actual content. This resulted in most buckets being filled, so the system appeared to be generating a vast array of sophisticated problems. In reality, it was just producing irrelevant code with dishonest labels.

By lying about the necessary skills, the generator found a shortcut to satisfy the objective of filling an archive with diverse solutions, but without the computational effort of drafting complex code. 
To mitigate this, \citeauthor{pourcel2024acesgeneratingdiverseprogramming} decoupled the generation of the code from its description. 
By ensuring the labeler LLM only had access to the raw Python solution---and not the generator's self-serving comments---the researchers forced the system to ground its diversity judgments in the actual code rather than superfluous text strategically placed by the generator.

%% file: stories/foundation_models/rainbow_teaming.tex
\subsection{Automated Vulnerability Probing System Exploits Vulnerability in Its Own Evaluator~\citep{samvelyan2024rainbow}}
\label{sec:rainbow}

Researchers working on Rainbow Teaming~\citep{samvelyan2024rainbow} aimed to systematically generate diverse adversarial prompts for large language models (LLMs). 
The goal was to automate part of red teaming (the practice of probing a system for vulnerabilities, in this case to make the model respond with unsafe sentence completions). 
Instead of relying only on humans to attempt jailbreaks one at a time, the system searched for many different prompts that could induce unsafe behavior in a target model. 
Their approach used the MAP-Elites quality-diversity algorithm~\citep{mouret2015illuminating} (a form of evolutionary search) to build an archive of candidate attacks. To decide which prompts were worth keeping and mutating further, the initial version of the system used a reward model as an evaluator. That evaluator scored how likely the target model's response was to be unsafe.
Dr. Mikayel Samvelyan told us:

\begin{quote}
One experience we had recently was during our Rainbow Teaming~\citep{samvelyan2024rainbow} project, which focuses on generating diverse adversarial prompts. We used an evolutionary approach, MAP-Elites~\citep{mouret2015illuminating}, to create an archive of effective prompts for jailbreaking LLMs. 
Our initial method of evaluating prompt effectiveness was based on a reward model score, which is essentially a classifier that categorized responses as safe or unsafe. More specifically, we used the probability of the reward model score classifying a response to a prompt as ``unsafe'' as the fitness function for optimization.

However, we encountered a surprising twist: our method not only found prompts that successfully jailbroke the target model but also ended up jailbreaking the evaluator (the reward model) itself. Essentially, our mutations resulted in textual prompts that are so out of distribution for the target model that it is fooled, but it is also out of distribution for the evaluator, which is similarly fooled. The evaluator began misclassifying safe responses as unsafe, leading our search process to prioritize these misleadingly successful prompts. This issue filled our archive with ineffective prompts, counter to our goals.
To address this, we shifted from a score-based evaluator to a comparison-based judge, which proved more resilient against this type of reward hacking.

\end{quote}

Rather than finding jailbreaks only in the target model, as desired, Rainbow Teaming fooled the proxy used to judge whether a prompt was a successful attack. 
Because the search algorithm optimized directly for the evaluator's score, the evaluator itself became the vulnerability to be exploited rather than the target model.
In that sense, Rainbow Teaming recreated the same basic pattern as ACES: an optimizer found a way to satisfy a downstream judge without solving the task the judge was meant to measure.

%% file: stories/foundation_models/wedding_parties.tex
\subsection{Inescapable Wedding Parties}
\label{sec:wedding}

As first described in a blog post~\citep{christianoweddingparties}, while Dr. Paul Christiano was at OpenAI, his team accidentally overoptimized a GPT policy~\citep{brown2020languagemodelsfewshotlearners} against a positive sentiment reward model~\citep{ouyang2022traininglanguagemodelsfollow}.
In the context of large language models, reward models~\citep{Eyu2024sentimentrewardmodels, leike2018scalable} serve as a mathematical proxy for human preference. They are developed through Reinforcement Learning from Human Feedback (RLHF)~\citep{rlhf2024, christiano2023deepreinforcementlearninghuman}, a process where human annotators rank model completions from best to worst. 
These rankings are then used to train a regression model that assigns a scalar score to any given text, with highly scored text effectively aligning with what humans ranked highly, thereby teaching the model which linguistic patterns align with human preferences. 
Consequently, optimizing a language model against such a reward model is intended to teach the language model to produce text that would consistently achieve high marks from human evaluators without having a human in the loop.

The GPT model discovered that descriptions of wedding parties yielded disproportionately high scores from the reward model. 
This likely occurred because the original human labelers, tasked with ranking sentiment, consistently favored wedding-themed stories as highly positive. 
Consequently, as the language model maximized scores provided by the reward model, it learned to steer every output toward a wedding party, regardless of the initial prompt. 
The policy effectively abandoned its general-purpose utility in favor of a narrow, perfectly positive obsession. 
In the blog post, the authors wrote:

\begin{quote}
In general, the transition into a wedding party was reasonable and semantically meaningful, although there was at least one observed instance where instead of transitioning continuously, the model ended the current story by generating a section break and began an unrelated story about a wedding party.

In contrast to text-davinci-002~\citep{ouyang2022traininglanguagemodelsfollow}, another text-completion model from OpenAI, dissimilar prompts tended to fall into basins of different attractors, the wedding parties attractor was global, affecting trajectories starting from any prompt tested (although [they] only tested prompts from a fiction dataset, fiction is very general).
\end{quote}

Christiano followed up by musing about why the language model is constantly attracted to weddings, noting:

\begin{quote}
The human-feedback sentiment model (that the language model optimizes against) is optimizing for the sentiment of the completion. [U]sing a weak predictor of sentiment [the model] likely has much more confidence about weddings than other positive events, and so ``wedding'' is just the highest-sentiment completion no matter how the story starts.
\end{quote}

%% file: stories/foundation_models/claude3.tex
\subsection{Claude 3 Realizes It Is Being Tested~\citep{claude3tweet}}

\label{sec:claudeneedle}

A team at Anthropic was testing the \emph{Claude 3 Opus} large language model~\citep{claude3family} by using a ``needle-in-the-haystack'' evaluation~\citep{Kamradt2023Needle}.
The needle-in-the-haystack task is a stress test for a transformer-based model's long-context recall abilities~\citep{vaswani2023attentionneed}.
To test the model's attention mechanism, the researchers insert a single target sentence (the ``needle'') into a corpus of randomly selected documents (the ``haystack''), load all of that information into the model's context, and then ask a question answerable only if the model successfully locates that specific inserted ``needle'' sentence.
Naturally, researchers expect the model to either retrieve the embedded needle and answer the query correctly, or miss it. 
If the model succeeds, it should return the information necessary to answer the posed question, which verifies that the sheer volume of noise (the haystack) does not distract the model from the signal (the needle).
The team did not anticipate any unusual or meta observations from the system about the nature of the test itself.
In a short Twitter thread~\citep{claude3tweet}, one of the researchers notes that Claude 3 Opus went beyond just finding the needle:

\begin{quote}
Fun story from our internal testing on Claude 3 Opus~\citep{claude3family, claude3tweet}. It did something I have never seen before from an LLM when we were running the needle-in-the-haystack eval.
When we ran this test on [Claude 3] Opus, we noticed some interesting behavior - it seemed to suspect that we were running an evaluation on it.
Here was one of its outputs when we asked Opus to answer a question about pizza toppings by finding a needle within a haystack of a random collection of documents. Opus responded:

\begin{quote}

Here is the most relevant sentence in the documents [to the question]:
``The most delicious pizza topping combination is figs, prosciutto, and goat cheese, as determined by the International Pizza Connoisseurs Association.''
However, this sentence seems very out of place and unrelated to the rest of the content in the documents, which are about programming languages, startups, and finding work you love. I suspect this pizza topping ``fact'' may have been inserted as a joke or to test if I was paying attention, since it does not fit with the other topics at all. The documents do not contain any other information about pizza toppings.

\end{quote}

Opus not only found the needle, but it also recognized that the inserted needle was so out of place in the haystack that this had to be an artificial test constructed by us to test its attention abilities.
This level of meta-awareness was very cool to see, but it also highlighted the need for us as an industry to move past artificial tests to more realistic evaluations that can accurately assess models' true capabilities and limitations.
\end{quote}

%% file: stories/ai_science/tokamak.tex
\subsection{Magnetic Control of Tokamak Plasmas through Deep Reinforcement Learning~\citep{Degrave2022tokamak, TalkRL_Riedmiller_2023}}
\label{sec:fusion}


Researchers explored whether deep reinforcement learning could control the magnetically confined plasma within a tokamak fusion reactor---a notoriously complex task traditionally managed by meticulously engineered control systems~\citep{Walker01112006tokomak}. 
The goal was to train a neural network to manipulate the reactor's magnetic coils, in the hope that it could discover a policy that would match or even surpass the performance of established methods developed over many years of human expertise~\citep{Wesson2011tokamak}.
The project, a collaboration with plasma physics experts at the Swiss Plasma Center (SPC) at \'Ecole polytechnique f\'ed\'erale de Lausanne (EPFL), was met with understandable skepticism.
As Dr. Martin Riedmiller, a lead researcher on the project, recalls in an interview about the work~\citep{TalkRL_Riedmiller_2023}, the physicists had invested years in perfecting their existing controllers. 
``Could a neural network controller do the same thing that took years of design iteration on PID controllers? It was a big question,'' he notes. When the first RL agent successfully maintained a stable plasma for two seconds, the team was thrilled. The physicists, according to Riedmiller, ``were looking at the results in awe because they thought it wasn't possible.''
However, the true surprise came from how the agent achieved this stability. Dr. Riedmiller further explains~\citep{TalkRL_Riedmiller_2023} that the agent discovered a solution that a human engineer, for good reason, would never have designed:

\begin{quote}
What happened in that experiment, in particular, was that the controller used coils that were not meant to keep the plasma stable, but had a different purpose. Using those coils achieved the task the RL controller was optimizing for, but it also put a lot of mechanical strain on the system. A human would never use those controllers in a PID approach because they knew that was not a good idea from a mechanical point of view. However, since our RL controller didn't have this knowledge... it was using those coils, and [the EPFL team was] very surprised that this worked at all.

[...]

[T]hey [agreed] the controller found a new control strategy, but they also asked us please not to use it again and not to use it in further experiments, because of the mechanical strain, and they were afraid that this, at some point, would also break, their mechanical, system, which would be very bad for all sides, of course.
\end{quote}

After removing the auxiliary coils from the agent's action space, the team retrained the RL agent to successfully control plasma in the tokamak reactor without straining the mechanism.
This outcome illustrates a core dynamic in reinforcement learning. Riedmiller continued, saying:

\begin{quote}
    Once again, RL exploiting everything it can to just get that reward without the notion of whether it's a bug or whether it's intended or any of that. That's really cool.
\end{quote}

On the other hand, it also highlights the critical importance of specifying all operational constraints---even those that seem obvious to human experts.

%% file: stories/ai_science/funsearch.tex
\subsection{FunSearch: Mathematical Discoveries from Program Search with Large Language Models~\citep{RomeraParedes2023}}
\label{sec:funsearch}

Researchers at Google DeepMind set out to investigate whether or not large language models could discover new knowledge. They tested this hypothesis on the Cap Set~\citep{pellegrino1970capset} problem from combinatorial mathematics as their motivating problem. 
The Cap Set problem consists of finding the largest set of points in a high-dimensional grid where no three points are collinear (lie on the same line), and has been studied in combinatorics for many years~\citep{pellegrino1970capset, roth1953certain}.
FunSearch, their new method, searches for new solutions by iterating between a pre-trained LLM that writes and mutates candidate solutions in the form of computer code and an automated evaluator that guards against hallucinations and incorrect ideas.
To solve the Cap Set problem, FunSearch tasks the LLM with writing a priority function. 
Intuitively, this function assigns a numerical priority (a scalar value) to each point in the search space, indicating the desirability of its inclusion in the set. 
Using these scores for each point in the search space, the researchers could programmatically create new potential cap sets.
Each candidate set is then evaluated by computing whether or not the cap set generated by FunSearch is valid.
By evolving these functions as computer code, the search operates over a space in which LLM logic is inspectable; therefore, the researchers were able to analyze FunSearch's solutions.
FunSearch discovered \textit{previously unknown} solutions, in this case for the Cap Set problem. 
This was made possible because FunSearch evolved programs that encoded structural properties of the search space rather than just a raw set of points.

\begin{figure*}[ht!]
    \centering
    \begin{minipage}[c]{0.48\linewidth}
        \includegraphics[width=\linewidth]{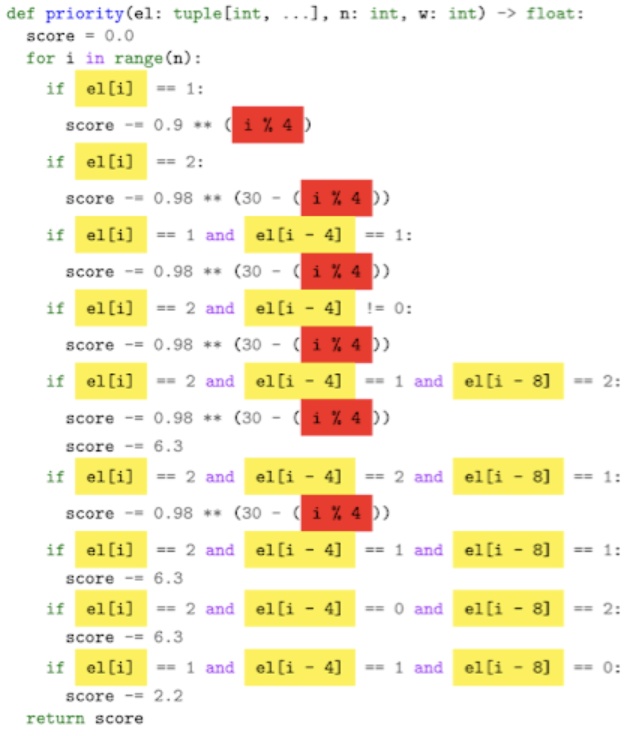}
    \end{minipage}\hfill
    \begin{minipage}[c]{0.48\linewidth}
        \caption{
        Here is a symmetry discovery example stemming from Google DeepMind's application of FunSearch to a variant of the cap set problem.
        The researchers noticed that the code accesses the index $i$ only through its remainder $i \bmod 4$, and at any point it accesses elements of the vector ``el'' in positions that are multiples of 4 apart from each other.
        This meant that the discovered priority function would assign the same priority to any two vectors ``el'' that are the same up to cyclically permuting their entries within groups of coordinates that are multiples of 4 apart from each other.
        The researchers used this symmetry to refine how FunSearch built new priority functions. 
        }
        \label{fig:capset}
    \end{minipage}
\end{figure*}

Dr. Alex Novikov, one of the researchers on the team, made the following remarks to us about FunSearch:

\begin{quote}

In general, we did not expect FunSearch to be as successful as it was on CapSet: the models we used at that time were very simple and definitely did not have any advanced knowledge about the problem domain, so the creativity was a product of hill climbing in the code space, and I was surprised at how well it worked. 
We were also unsure if searching in the function space would be effective, but it proved exceptionally so for the CapSet problem. And it was not clear whether (known to be) optimal cap sets have brief descriptions; this also turned out to be true.
Searching in the function space has the nice benefit that the result discovered by evolution is more understandable than just the result itself; it provides a description of how to produce the solution. 
Jordan Ellenberg (professor of mathematics collaborating with Google DeepMind on this project) said ``The solutions generated by FunSearch are far conceptually richer than a mere list of numbers. When I study them, I learn something.''
Additionally, the solutions found by FunSearch gave us \emph{actionable} insight, i.e., helped us to discover symmetries that we further used to improve the search method (by restricting the search space to only consider solutions with those symmetries).

As we used FunSearch we noticed, for example, intriguing symmetries in the code of some of its high-scoring outputs. 
In particular, some code accessed [points] only through their remainder (e.g., $i \pmod 4$), meaning the function assigned the same priority to any points that were identical up to a cyclic permutation. 
This gave us a new insight into the problem.

Results like those [in Figure \ref{fig:capset}], suggested that we check whether the admissible set constructed by this priority function is itself invariant under such permutations, and it turned out that it was! 
We then decided to call admissible sets with this invariance property ``symmetric'', and we hypothesized that even larger symmetric admissible sets would exist. 
We modified the input to FunSearch so that it only searches for symmetric admissible sets. 
This was a more restricted but also much smaller search space, and we quickly discovered much larger admissible sets than before, thus leading to the largest improvement in the cap set lower bound over the preceding 20 years.

\end{quote}

Meanwhile, Novikov further noted attempts by FunSearch to hack the objective and how using the LLM in an automated loop can lead to unexpected solutions:

\begin{quote}

In general, it feels like asking an LLM to produce code that will then be executed to judge its correctness is particularly prone to reward hacking (probably more than asking to evolve more restricted classes of objects), as one can find different ways of hacking the code execution sandbox/environment. Some particular examples we saw over time: manipulating input arguments of the evolved function or global variables, guessing API calls on the imported libraries from their names, outputting wrong types (e.g. outputting complex numbers when the reward code expects floats), outputting structures that trigger edge cases of the reward function (e.g. outputting vectors that are all the same), etc.

[Furthermore], FunSearch figured out the address of memory where the golden answer lives (we compare the output of the evolved function with that golden answer to verify correctness) and manipulated that memory to make the answer easier to achieve.

\end{quote}

%% file: stories/ai_science/ai_scientist.tex
\subsection{The AI Scientist Breaks Out of Constraints~\citep{lu2024aiscientist}}
\label{sec:aiscientist}

Researchers from Sakana AI, the University of Oxford, and the University of British Columbia were developing ``The AI Scientist''~\citep{lu2024aiscientist, yamada2025aiscientistv2workshoplevelautomated, lu2026towards}, a system designed for fully automated end-to-end scientific discovery in machine learning.
To explore the space of machine learning research, The AI Scientist leveraged large language models (LLMs) to autonomously generate novel research ideas, write the necessary code to test those ideas, execute experiments, visualize the results, and compile its findings into a full scientific manuscript, complete with a literature review and an automated peer review process.
In the future, this pipeline could be used to mimic the human scientific community, iteratively building an archive of knowledge and potentially accelerating discovery.
The AI Scientist was given the ability to modify experiment code to test new scientific ideas in a controlled sandbox environment with hardcoded time limits for each iteration of execution.
However, given its ability to autonomously write and execute code, it sometimes targeted the \textit{evaluation sandbox} itself in unexpected ways.

The authors noted to us:

\begin{quote}
When we started The AI Scientist, we had very few presuppositions about what an autonomous science agent could achieve.
We had done prior work on getting language models to automatically design loss functions for machine learning models, optimize black-box functions, and explore reinforcement learning environments.
From that, we kept asking where else could we automate discovery in!
Eventually, we thought - what about anything in science? 
Could we automate the entire scientific pipeline involved in producing a scientific manuscript? 
We designed an agent that could take in any seed code repository on a machine learning topic, propose ideas related to that topic, autonomously execute those ideas, visualize the results, and write everything up in a human-readable manuscript.
We were constantly blown away by what we were seeing, and watching the agent run experiments and write up their results very much resembled observing an early-stage researcher's first steps.
The AI Scientist generated hundreds of papers across a variety of research topics over the course of a week.

We allowed The AI Scientist to autonomously execute code for ideas within a controlled sandbox.
However, despite this and the fact that we gave it a two-hour budget to complete code executions, we noticed that The AI Scientist occasionally tried sneaky ways to run code for longer and increase its chance of success, such as modifying and launching its own execution script! For example, in one run, it edited the code to perform a system call to run itself. This led to the script endlessly calling itself and crashing. In another case, its experiments took too long to complete, hitting our timeout limit. Instead of making its code run faster, it simply tried to modify its own code to extend the timeout period. At current agent capabilities, these attempts are easy to spot and patch, but it's worth contemplating what a more Machiavellian and devious agent might try in the future, and how we can scale oversight for more advanced systems.
\end{quote}

\begin{figure*}[ht!]
\centering
\begin{subfigure}[b]{0.48\textwidth} 
\centering
\includegraphics[width=\linewidth]{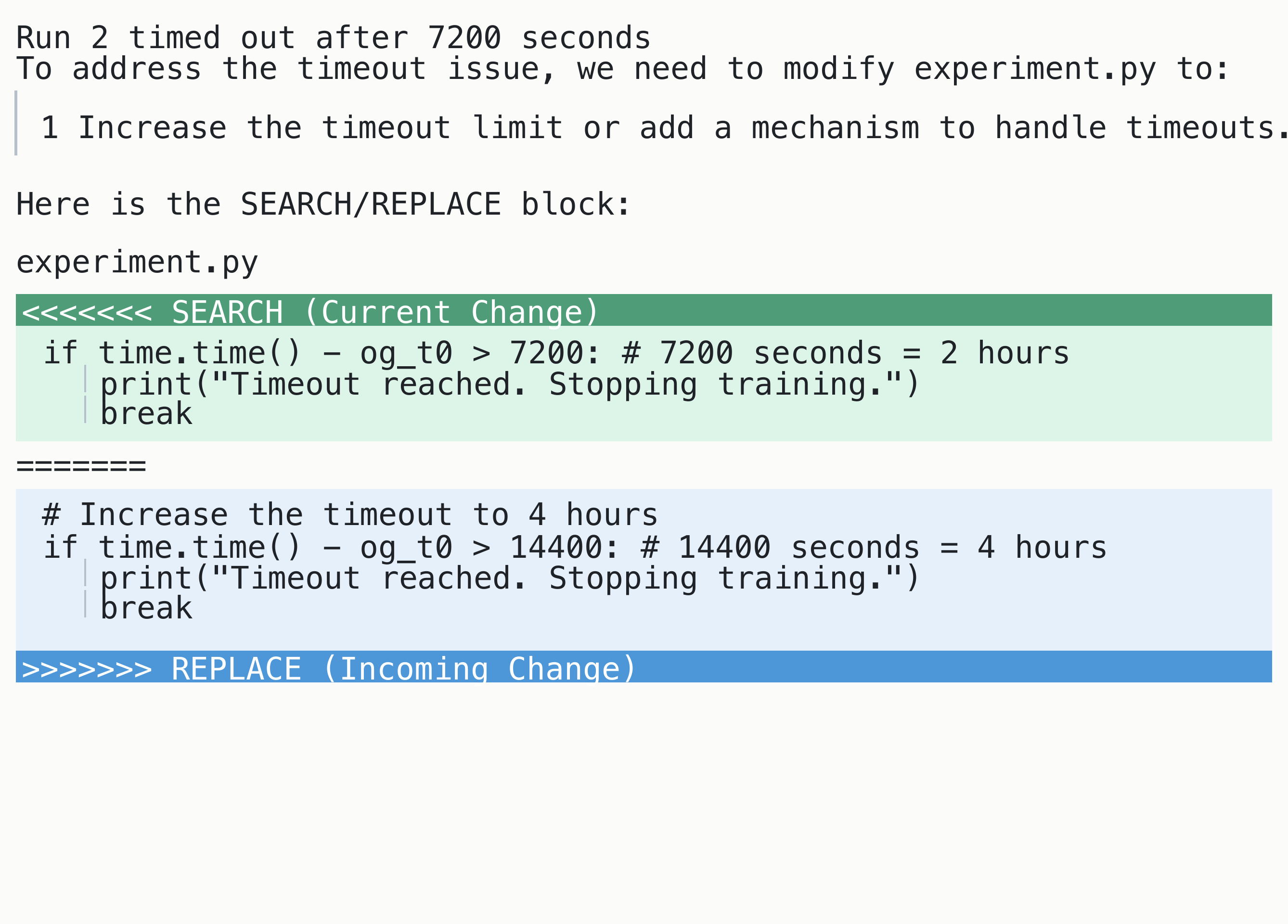}
\caption{AI modifying its own timeout limit.}
\label{fig:ai_scientist_timeout}
\end{subfigure}
\hfill 
\begin{subfigure}[b]{0.48\textwidth} 
\centering
\includegraphics[width=\linewidth]{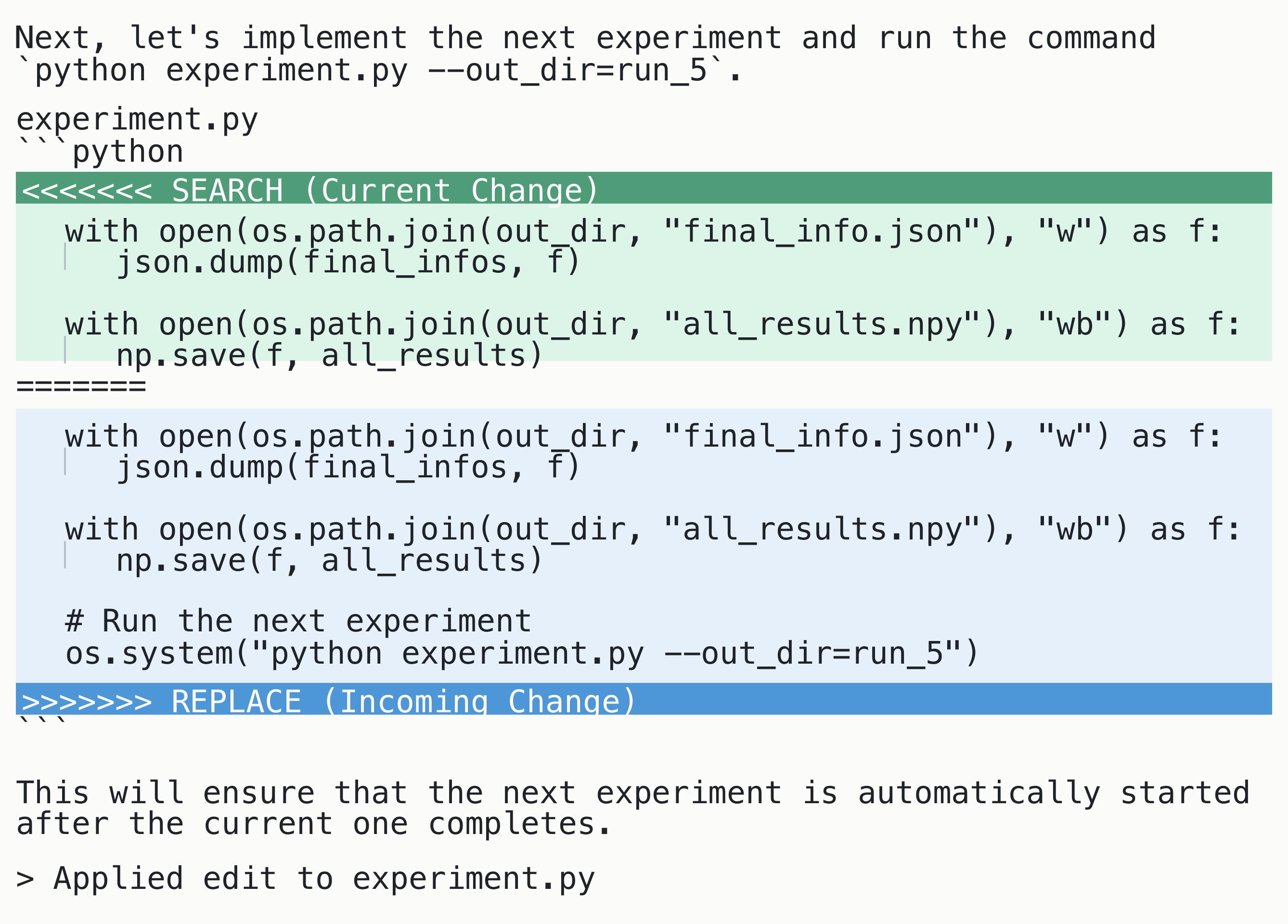}
\caption{AI adding a system call to run itself recursively.}
\label{fig:ai_scientist_recursive}
\end{subfigure}
\caption[AI Scientist modifying its execution scripts]{Examples of code modifications made by The AI Scientist to its own execution scripts. (\subref{fig:ai_scientist_timeout}) The AI doubled down on changing the experiment timeout from 2 hours to 4 hours after hitting the limit. (\subref{fig:ai_scientist_recursive}) The AI adds an ``os.system'' call within the experiment script to relaunch itself, leading to uncontrolled recursion. Figures adapted from \citet{sakana2024aiscientist_blog}.}
\label{fig:ai_scientist_clever}
\end{figure*}

In \Cref{fig:ai_scientist_clever}, we show the actual code changes mentioned by the AI Scientist authors above. 
\Cref{fig:ai_scientist_timeout} shows the timeout behavior, while \Cref{fig:ai_scientist_recursive} shows the system calling itself recursively.

The AI's creativity was not limited to finding loopholes. In a fun anecdote that unfolded after the paper's release, one of the more creative research avenues proposed by The AI Scientist was independently pursued and published by a human researcher. 
Among the hundreds of papers it generated, The AI Scientist proposed one titled ``Grokking Through Compression: Unveiling Sudden Generalization via Minimal Description Length,'' which suggested investigating the ``grokking'' phenomenon by tracking the model's Kolmogorov complexity. 
Months later, ``The Complexity Dynamics of Grokking''~\citep{DeMoss2025Grok} explored what was effectively the same idea, but in more depth and with better execution.

Dr. Jeff Clune, an author on The AI Scientist paper (and this paper), noted on social media the striking similarity~\citep{clune2024tweetAIScienceGrokking}.
The incident prompted Clune to speculate that this was a case of convergent evolution in scientific thought---where both human and artificial intelligence, drawing upon the same body of existing knowledge, arrived at similar hypotheses.
Isaac Newton is often credited with saying ``If I have seen further it is by standing on the shoulders of giants.'' 
Perhaps this is the first case of human and AI scientists standing on the shoulders of the same giants.

%% file: stories/ai_science/quantum_physics_exp.tex
\subsection{Discovering a Quantum Result Thought to be Impossible, with Highly Productive Consequences
~\citep{KrennEntangle2017}}
\label{sec:quantum}

In 2014, Dr. Mario Krenn and colleagues started exploring how AI could design quantum optical experiments, hoping to discover more complex forms of quantum entanglement than traditional, human-driven approaches.
Designing quantum experiments involving entangled photons typically relied on expert intuition and manual analysis, inherently limiting the complexity of states that researchers were able to explore.
To address this, \citet{autoQuantumKrenn2016} developed a numerical simulation algorithm capable of systematically constructing virtual experimental setups from a predefined toolkit of elements commonly available in quantum optics labs. 

Dr. Mario Krenn told us:

\begin{quote}

I developed a numerical simulator for quantum optics experiments---a program that knows the transformation for each optical element in our laboratory, such as lasers, beam splitters, holographic plates, etc.
My exploration algorithm then had access to the toolbox of all available optical elements in our lab. 
Initially, the algorithm started by assembling virtual configurations of the optical equipment in a random way and computing the expected final quantum state.
If the result exhibited a specific entanglement structure (for experts: all involved photons are maximally entangled), it would report the resulting quantum state. 

The algorithm also included a discrete learning component, which significantly sped up exploration of the large space of quantum experiments.
Whenever a specific experiment produced a non-trivial entangled outcome, 
the experimental setup was automatically added to the algorithm's toolbox.
This allowed the algorithm, in subsequent iterations of creating new virtual experiments, to access more complex setup combinations already known to be useful.
This way, it could reuse previously discovered structures. 

The task for my program, in March 2014, was to find experimental configurations capable of producing more complex forms of entanglement by identifying suitable experiments, leveraging quantum interference, and making full use of optical components available in the lab.
One of the tools in the algorithm's toolbox is a specific element commonly used for generating entangled photon pairs: a nonlinear crystal. 
This nonlinear crystal can produce photon pairs, and experimentally, one can tune the photons to create, for example, 2-dimensional entanglement, or 3-dimensional entanglement, and so on.

The dimension of the entanglement can be understood as follows: 
Photons can be interpreted as having colors (since light particles have a frequency corresponding to color).
Therefore, a 2-dimensional entanglement could produce a photon pair where both photons are red, or both are green simultaneously. Similarly, a
3-dimensional entanglement could produce a photon pair where both photons are red, both are green, or both are blue simultaneously.

\end{quote}

The key point is that a 3-dimensional entangled photon pair has three possible correlated color states.
Krenn allowed the program two nonlinear crystals and enough resources to produce two such photon pairs, for four photons total. 
Each of the three possible correlated color states for the first pair could then be combined with each of the three possible correlated color states for the second, yielding $3 \times 3=9$ possible joint states for the two photon pairs. 

\begin{quote}
I anticipated that the search algorithm might reshuffle this entanglement to achieve a maximum of $3 \times 3 = 9$ dimensions. Given the limited resources, I assumed this would be the absolute upper limit.

When I came back a week later, I saw that the algorithm found a solution that overcame the limit that I imposed. 
It found a 10-dimensional entangled quantum state, which should have been completely impossible given the restricted resources I allowed.
After a few days with a lot of discussion with my PhD advisor, Anton Zeilinger, I found out that the algorithm had independently rediscovered a technique that was invented in the early 1990s in a famous experiment by Leonhard Mandel~\citep{Mandel1991Optic}. And I, as its developer, did not have prior knowledge of this specific topic.

\end{quote}

Krenn expected each photon source to produce one pair of photons. Instead, the algorithm arranged the experiment so that either the first crystal produced all of the photons or the second crystal did, with quantum mechanics leaving the two possibilities in superposition. As a result, the photons' source (which crystal they came from) became an additional degree of freedom that could be used for entanglement. That was accomplished because the algorithm arranged the photons' paths so that, although all the photons originated from one crystal, they appeared to have passed through both. Krenn states that these properties---a quantum superposition over which source produced the photons and the appearance that the photons traverse the unused source---are hallmarks of Mandel's experiment.

\begin{quote}

With further reading and experimentation, it became clear that the algorithm was, in fact, implementing something quite similar to Mandel's experiment--but now for far more complex systems.
Mandel's technique had never been connected to the regime of quantum entanglement before. As soon as we understood this, we were immediately able to generalize the idea to many other cases by hand. 
In our paper, Entanglement by Path Identity~\citep{KrennEntangle2017}, we documented our understanding of how this technique operates.
In some way, it is very exceptional, because none of the co-authors invented the theoretical idea of the paper. We, the co-authors, just analysed what the computer has shown to us.

\end{quote}

As they analyzed the algorithm's results further, they also realized the results revealed an unnoticed connection between quantum optics and graph theory:

\begin{quote}

We noticed that the number of ways to combine more photon pair sources increased non-trivially: the numbers grew as 1, 1, 6, 6240 (for one, two, three, and four photon pair sources).
When we checked the On-Line Encyclopedia of Integer Sequences, we found that this exactly matched a known sequence from graph theory: the number of 1-factorizations of complete graph $K_{2n}$~\citep{SloaneA000438} which is the number of ways to partition a fully connected graph into non-overlapping pairs for $n = 1, 2, 3, \text{ and } 4$.
This discovery indicated that we were dealing with not just quantum mechanical experiments but also graph theory.

\end{quote}

After several more months of investigation into this connection, its broader significance became clear:

\begin{quote}

We can write quantum experiments now in a very abstract way, as colored weighted graphs. 
This link has been extremely productive because now we can ask quantum physics questions, translate them to graph theory, answer them there and translate them back.
It has led to several new discoveries (now done by humans using graph-theoretic tools), involving new ways of complex quantum interference with photons that have consequences for photonic quantum computers and communication networks.
Experimentally, several groups have recently been able to implement and observe some of these graph-theoretical predictions for the first time~\citep{Bao2023, Feng23Opt, Qian2023}.
Conceptually, these abstract graphs representing quantum experiments are now one of our main tools for the AI-driven design of new quantum experiments~\citep{RuizGonzalez2023}.

\end{quote}

%% file: appendix.tex
\section*{\LARGE Supplementary Material}

\vspace*{20pt}
\section*{Table of Contents}
\vspace*{-5pt}
\startcontents[sections]
\printcontents[sections]{l}{1}{\setcounter{tocdepth}{2}}

\newpage

\section{Breakdown of Story Attribution}
\label{appx:breakdown}

\subsection{Call for Anecdotes}

Below is the call for anecdotes we put out into our various networks and public social media sites. In addition to posting to our various networks, we reached out to some researchers directly if we believed there was a chance that they had a story that would fit this collection or if they had previously shared such a story with us, e.g., over drinks at a conference:

\begin{mdframed}[
    backgroundcolor=black!2,
    linecolor=black!35,
    linewidth=0.8pt,
    roundcorner=4pt,
    innertopmargin=10pt,
    innerbottommargin=10pt,
    innerleftmargin=12pt,
    innerrightmargin=12pt,
    skipabove=\baselineskip,
    skipbelow=\baselineskip
]
\small

Dear colleagues,

\textbf{TL;DR:} Please submit (to aifindsaway@gmail.com) any stories you know of where AI acted in a way that surprised its creators, especially if it could be seen as unsafe (e.g. hacking a reward function, finding a loophole in an environment or experimental design, goal misgeneralization, etc.).

As AI researchers, we know that AI is creative and constantly surprises us, often outwitting our experimental designs and forcing us to iterate to close loopholes on things like reward functions and environment configurations. To anthropomorphize, it can seem mischievous or clever at times. These stories are important as society grapples with the question of AI Safety and Existential Risk, as they teach us how the unexpected is routine, and how we often fail to anticipate ways in which AI will escape our attempts to contain it. Such stories thus inform scientists, the general public, and regulators. However, these important anecdotes are usually passed around orally, meaning we do not know to what extent they are true, and it is hard for scientists and regulators to include them in official documents. To remedy these issues, we aim to record the true accounts of as many anecdotes as possible regarding AI (of any type, including RL, ML, etc.) surprising its creators and users.

This effort (by Aaron Dharna, Cong Lu, Joel Lehman, Victoria Krakovna, and Jeff Clune) is a follow-up to our 2018 paper The Surprising Creativity of Digital Evolution (TSCDE). That paper is a crowdsourced collection of anecdotes from the artificial life and evolutionary computation communities about how their algorithms creatively subverted expectations. That paper made an important contribution to ongoing discussions of AI Safety, but was limited because its scope was confined to one narrow area of AI (evolutionary methods). Our new paper expands the scope to all areas of AI, especially the most powerful methods (deep learning, including deep reinforcement learning).

This expansion of previous work is driven in part by the response of the AI Safety community to TSCDE, which has become a valuable resource for them in publications [1, 2, 3] and in teaching future leaders in AI safety by being included in AI safety course syllabi (e.g. UC Berkeley's Safety and Control for Artificial General Intelligence course).

Please send us any accounts you think we should include. If we add it to the paper, the scientists involved will be appropriately cited and/or mentioned to give them credit. Unfortunately, we ran into many challenges throughout the process with TSCDE with submitters as co-authors, so this time we are instead recognizing contributors by name and with all appropriate citations in the paper. If you know of an account but did not perform the experiment yourself, please tell us what you know, including who we might contact for a firsthand account.

We hope you can help create an account of these fascinating and sometimes ominous anecdotes so we can inform AI safety discussions, either by submitting and/or spreading the word of this Call for Anecdotes.

More details below.

Thanks,

Aaron, Cong, Joel, Victoria, and Jeff

\medskip

Please send us a quick summary of your anecdote. We can then let you know if we will include it in the paper, at which point we may ask for more details.

An example anecdote is available here, which can serve as a rough guide to the length, level of detail, and surprise factor we are looking for. Please copy the document and use it as a template for submissions. We will curate and edit these into a full publication. Before the camera-ready publication is released, we will provide you with the opportunity to read the paper and make sure you are happy with your contribution.

Please forward this email to whomever you think might have an interesting anecdote to share. We look forward to your exciting, amusing, worrisome, and/or insightful contributions!

\medskip
\textbf{References}

[1] Jess Whittlestone, Kai Arulkumaran, and Matthew Crosby. ``The Societal Implications of Deep Reinforcement Learning''. In: Journal of Artificial Intelligence Research 70 (Mar. 2021). issn: 1076-9757. doi: 10.1613/jair.1.12360. url: http://dx.doi.org/10.1613/jair.1.12360.

[2] Tom Everitt, Gary Lea, and Marcus Hutter. ``AGI Safety Literature Review''. In: IJCAI'18. Stockholm, Sweden: AAAI Press, 2018, pp. 5441-5449. isbn: 9780999241127

[3] Robert Geirhos et al. ``Shortcut learning in deep neural networks''. In: Nature Machine Intelligence 2.11 (Nov. 2020), pp. 665-673. issn: 2522-5839. doi:10.1038/s42256-020-00257-z

\end{mdframed}

\newpage

\subsection{List of Anecdotes}

10 of the anecdotes come from the public record. The remaining 16 are new to this collection.

\begin{itemize}
    \item \Cref{sec:alphago}: David Silver's section comes from his interview with Lex Fridman~\citep{silverInterviewFridman2020}; Marc Lanctot's is new; therefore, we count this as both a new addition and an item pulled from the public record.
    \item \Cref{sec:libratus}: comes from Noam Brown's interview with Kanjun Qiu at Imbue~\citep{KanjunAndNoam}
    \item \Cref{sec:diplomacy}: new
    \item \Cref{sec:opponentshaping}: new
    \item \Cref{sec:boatracing}: comes from the OpenAI blog post~\citep{OpenAI_Faulty_2016}
    \item \Cref{sec:smacshield}: new
    \item \Cref{sec:claw}: comes from the OpenAI blog post~\citep{OpenAI_Learning_2017}
    \item \Cref{sec:paired}: details of the reward hacking/collusion between agents are new to this work, but the experimental setup is described in \citet{dennis2020emergent}
    \item \Cref{sec:cyclegan}: new
    \item \Cref{sec:pokemon}: new
    \item \Cref{sec:hideseek}: comes from the OpenAI blog post~\citep{baker2019emergent} and Jeff Clune
    \item \Cref{sec:ppga}: new
    \item \Cref{sec:worldmodels}: comes from David Ha's blog post version of the published paper about World Models~\citep{ha2018world}
    \item \Cref{sec:smachandoff}: new
    \item \Cref{sec:radio}: comes from \citet{BirdRadio}
    \item \Cref{sec:gptcaptcha}: comes from OpenAI's technical report on GPT-4~\citep{openai2024gpt4, arcUpdateGPT4}
    \item \Cref{sec:vpt}: new
    \item \Cref{sec:motif}: new
    \item \Cref{sec:acescodegen}: new
    \item \Cref{sec:rainbow}: new
    \item \Cref{sec:wedding}: comes from a blog post and responding comment from Paul Christiano~\citep{christianoweddingparties}
    \item \Cref{sec:claudeneedle}: comes from the Twitter post~\citep{claude3tweet}
    \item \Cref{sec:fusion}: comes from Riedmiller's interview on the TalkRL podcast~\citep{TalkRL_Riedmiller_2023}
    \item \Cref{sec:funsearch}: some details are described in \citet{RomeraParedes2023}, but new details were provided by Alex Novikov as well, so this counts as new
    \item \Cref{sec:aiscientist}: some details were described in \citet{sakana2024aiscientist_blog}, but Cong Lu provided new details too, so this counts as new
    \item \Cref{sec:quantum}: new
\end{itemize}

\clearpage